\documentclass[10pt,twocolumn,letterpaper]{article}

\usepackage{wacv}
\makeatletter
\@namedef{ver@everyshi.sty}{}

\usepackage{times}
\usepackage{epsfig}
\usepackage{graphicx}
\usepackage{amsmath}
\usepackage{amssymb}
\usepackage{multirow}
\usepackage{caption}
\usepackage{subcaption}
\usepackage{pgffor}
\usepackage{pifont}
\usepackage[symbol]{footmisc}

\def\wacvPaperID{1006} 

 \wacvfinalcopy 

\ifwacvfinal
\def\assignedStartPage{1} 
\fi

\ifwacvfinal
\usepackage[breaklinks=true,bookmarks=false]{hyperref}
\else
\usepackage[pagebackref=true,breaklinks=true,colorlinks,bookmarks=false]{hyperref}
\fi

\ifwacvfinal
\else
\fi

\newcommand{\xmark}{\ding{55}}%
\newcommand{\cmark}{\ding{51}}%

\begin{document}

\title{Recursive Contour-Saliency Blending Network for Accurate Salient Object Detection}

\author{Yun Yi Ke, Takahiro Tsubono\thanks{Corresponding author}\\
AI Technology Lab, OPEN8 Singapore\\
{\tt\small \{yikey, tsubonot\}@open8.com}
}

\maketitle

\begin{abstract}
Contour information plays a vital role in salient object detection. However, excessive false positives remain in predictions from existing contour-based models due to insufficient contour-saliency fusion. In this work, we designed a network for better edge quality in salient object detection. We proposed a contour-saliency blending module to exchange information between contour and saliency. We adopted recursive CNN to increase contour-saliency fusion while keeping the total trainable parameters the same. Furthermore, we designed a stage-wise feature extraction module to help the model pick up the most helpful features from previous intermediate saliency predictions. Besides, we proposed two new loss functions, namely Dual Confinement Loss and Confidence Loss, for our model to generate better boundary predictions. Evaluation results on five common benchmark datasets reveal that our model achieves competitive state-of-the-art performance. 
\end{abstract}

\section{Introduction}
Salient object detection (SOD) aims to detect and segment the most attention-attractive region or object in a visual scene. Unlike eye fixation prediction (FP) \cite{TREISMAN198097}, SOD requires obtaining the entire region with clear boundaries. Due to its essential role and wide applications in image understanding \cite{IMU}, image captioning \cite{fang2015captions}\cite{xu2016show}, and video summarization \cite{video_summarization}, recently, various methods have been proposed in the field.

Since 2015, Convolutional Neural Networks (CNNs) \cite{ResNet}\cite{ImageNet} have been adopted for SOD tasks. Though algorithms like PiCANet \cite{PICANET}, BMPM \cite{BMPM}, and PAGRN \cite{PAGRN} achieved significantly better results, the predicted object usually has poor boundaries. To obtain more precise boundaries, proposed by Qin \textit{et al}., BASNet \cite{BASNET} adopted a boundary refinement U-Net \cite{UNET} at the end of the saliency detection network and trained their model using various losses. In \cite{CTLoss}, Chen \textit{et al}. proposed Contour Loss (CTLoss), which was a weighted Binary Cross-Entropy (BCE) loss, to improve the boundary predictions. Alternatively, models like EGNet \cite{EGNET}, PoolNet \cite{POOLNET}, and ITSD \cite{ITSD} fused contour predictions with saliency by explicitly supervising a contour branch. With contour cues, models yielded better boundary predictions.

\begin{figure}[!t]
\centering
\begin{subfigure}{.24\linewidth}
\centering
\includegraphics[width=\linewidth]{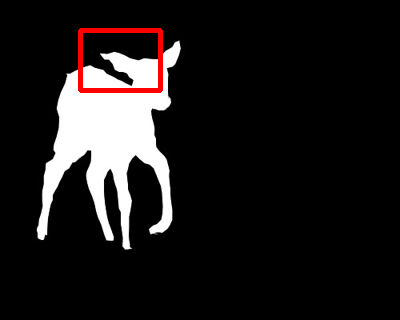}
\includegraphics[width=\linewidth]{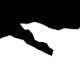}
\includegraphics[width=\linewidth]{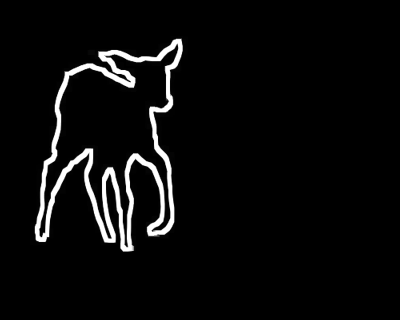}
\caption{}
\label{fig:gt_sal}
\end{subfigure}
\begin{subfigure}{.24\linewidth}
\centering
\includegraphics[width=\linewidth]{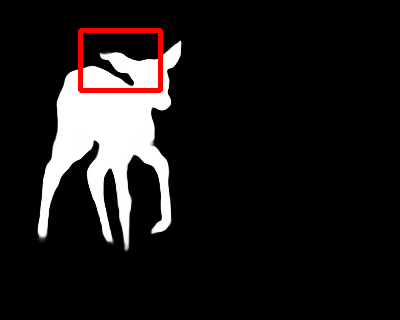}
\includegraphics[width=\linewidth]{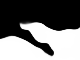}
\includegraphics[width=\linewidth]{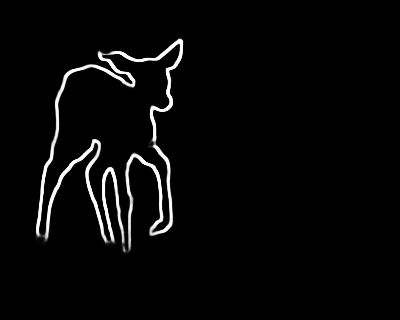}
\caption{}
\label{fig:sal_pred}
\end{subfigure}
\begin{subfigure}{.24\linewidth}
\centering
\includegraphics[width=\linewidth]{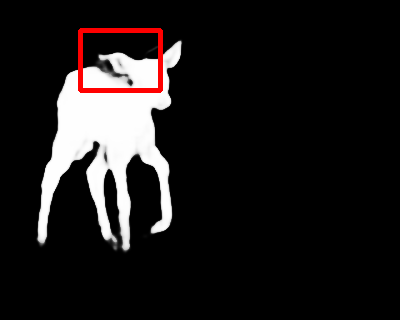}
\includegraphics[width=\linewidth]{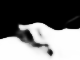}
\includegraphics[width=\linewidth]{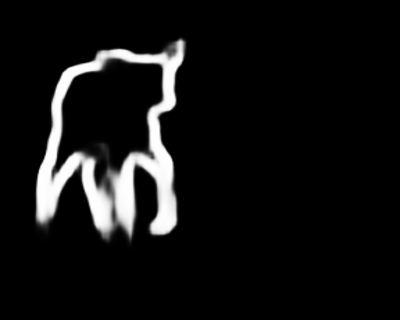}
\caption{}
\label{fig:itsd_sal}
\end{subfigure}
\begin{subfigure}{.24\linewidth}
\centering
\includegraphics[width=\linewidth]{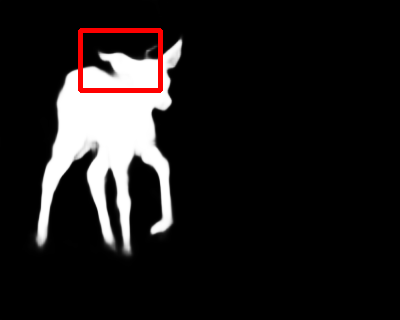}
\includegraphics[width=\linewidth]{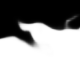}
\includegraphics[width=\linewidth]{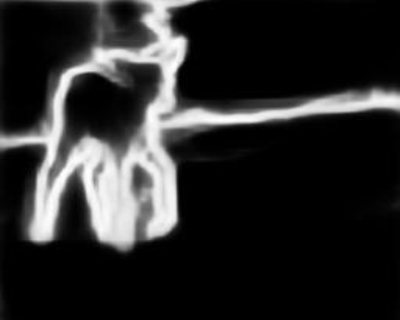}
\caption{}
\label{fig:pool_sal}
\end{subfigure}
\caption{Visual comparisons between contour-based models. Saliency (first row) and corresponding contour predictions (last row) are listed. Ground truth contours are obtained via erosion and dilation with kernel size of 5. (a) ground truth, (b) ours, (c) ITSD \cite{ITSD} and (d) PoolNet \cite{POOLNET}. }
  \label{fig:excessiveFP}
\end{figure}

\begin{figure}
  \centering
  \includegraphics[width=0.95\linewidth]{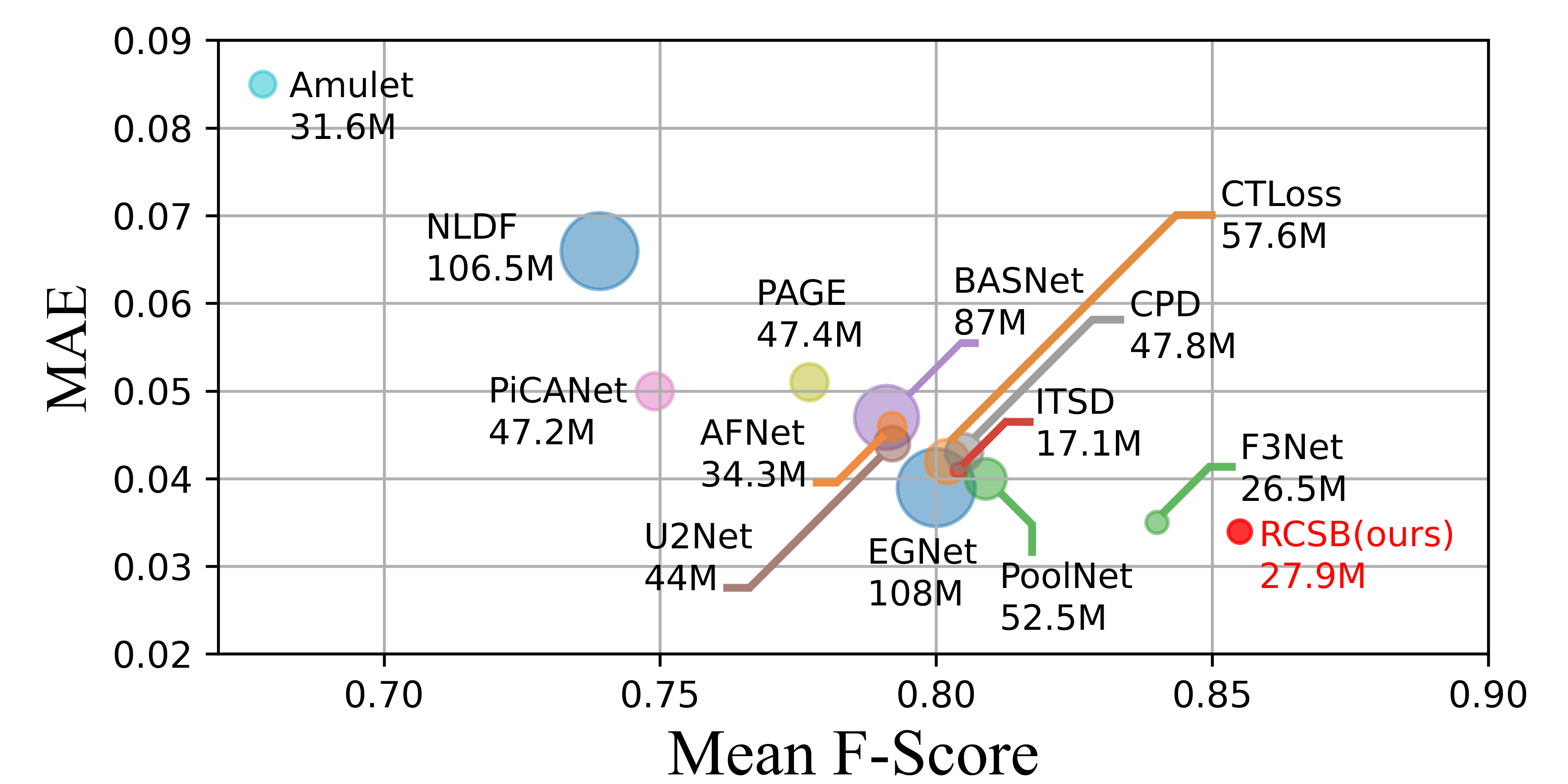}
  \captionof{figure}{Mean F-score, total parameter and MAE comparison between RCSB with 13 state-of-the-art models, including EGNet \cite{EGNET}, PiCANet \cite{PICANET}, C2SNet \cite{C2SNET}, PoolNet \cite{POOLNET}, PAGE \cite{PAGE}, Amulet \cite{AMULET}, ITSD \cite{ITSD}, BASNet \cite{BASNET}, U2Net \cite{U2NET}, NLDF \cite{NLDF},  AFNet \cite{AFNET}, F3Net \cite{F3NET} and CPD \cite{CPD} on DUTS-TE \cite{DUTSTE} dataset. Bubble size represents model size.}
  \label{fig:FPS}
\end{figure}

However, above mentioned models still hold several problems that can be further improved. First, for better performance, many studies have introduced a huge number of trainable parameters. EGNet contains 108 million parameters, BASNet and its extended work, U2Net \cite{U2NET}, have more than 87 and 44 million parameters, respectively (Fig. \ref{fig:FPS}). The huge number of parameters not only leads to increased consumption of computational resource, but also makes the model difficult to train. Second, though object boundary is greatly improved for contour-based networks  \cite{POOLNET}\cite{EGNET}\cite{ITSD}, predictions still have excessive false positives, as shown in Fig. \ref{fig:itsd_sal} and \ref{fig:pool_sal}. Finally, to our best knowledge, for all deep learning-based models in the current field, intermediate saliency or contour predictions are generated and supervised via side branches, which introduce redundant parameters and inefficiency.

To address the abovementioned issues, we proposed a recursive contour-saliency blending network, namely, RCSBNet, for high accuracy salient object detection. We adopted a recursive CNN to reduce total trainable parameters while we can make our model deep. Unlike previous studies \cite{POOLNET}\cite{EGNET}\cite{ITSD}, where the contour and saliency are explicitly trained via two branches, we introduced a Contour-Saliency Blending (CSB) module in our network so that contour and saliency are intertwined and fused every step in the recursion. Meanwhile, to further improve the efficiency, we proposed a Stage-wise Feature Extraction (SFE) module to directly supervise intermediate saliency and contour predictions in the primary network without using any side branch. Lastly, we divided the training task into accuracy and confidence, and proposed Dual Confinement Loss (DCLoss) and Confidence Loss (CLoss) respectively for better model performance. To sum up, our contributions are as follows:

(1) We proposed an efficient and accurate network, RCSBNet. By using a recursive CNN and the proposed Stage-wise Feature Extraction (SFE) module, contour and saliency are fused more efficiently and effectively.

(2) We developed two loss functions, DCLoss and CLoss, to further help the boundary prediction.

(3) Our model has only 27.9 million parameters, which is significantly smaller and efficient than most of the networks in the field (Fig. \ref{fig:FPS}).

(4) We conducted comprehensive evaluations on 5 widely used benchmark datasets and compared with 13 state-of-the-art methods. Our method achieves competitive state-of-the-art results among all datasets.

\section{Related Work}
Early approaches based on hand-crafted priors \cite{GCSOD}\cite{SODMARKOV}\cite{DUTO} have limited effectiveness and generalization ability. The very first deep salient object detection (SOD) methods \cite{HKUIS}\cite{SODCONTEXT} used multi-layer perceptron to predict saliency score for each image.  These methods suffered from low efficiency and damage of feature structures due to flattening. Later, some studies introduced a fully convolutional network (FCN) and achieved promising results.

\textbf{Recurrent Networks}. 
In \cite{RCNN}, a recurrent convolutional neural network (RCNN) was proposed for object detection. The main idea was to unfold the same convolution layer several times while weights are shared. It had the advantage that model depth can now be deeper by unfolding, while the total number of trainable parameters remains the same. It also revealed that, by increasing the number of recursions, better results would be obtained. 

In 2016, the recurrent CNN was introduced to salient object detection task, and proposed by Wang \textit{et al}. RFCN \cite{RFCN} recursively refines the saliency prediction from previous time step. Later, proposed by Kuen \textit{et al}. \cite{RASA}, a recurrent network was designed to refine selected image sub-regions iteratively. In \cite{PAGRN}, Zhang \textit{et al}. designed a multi-path recurrent model for saliency detection by transferring global information from deep layers to shallower layers. Hu \textit{et al}.  \cite{RADF} proposed their salient object detection model by concatenating multi-layer deep features recurrently. It was proved that saliency predictions will be refined by using recurrent mechanism.

\textbf{Utilizing Contour Information}.
In recent years some studies explored and verified the effectiveness of involving contour information to improve the accuracy of saliency prediction. In \cite{CTLoss}, Chen \textit{et al}. considered boundary pixels as hard samples and proposed a contour loss, which was a weighted BCE loss, to train their network. Qin \textit{et al}. \cite{BASNET} combined Structural Similarity Index (SSIM), Intersection over Union (IOU), and BCE as their contour-aware loss function to achieve better boundary quality. In another seminal work, Salient Edge Detector (SED) \cite{PAGE} was introduced to simultaneously generate saliency and contour predictions by using a residual structure. Furthermore, PoolNet \cite{POOLNET} applied multi-task training and fused the contour information with saliency predictions. Later, ITSD \cite{ITSD} proposed a two-stream network to convert saliency and contour interactively and yielded good boundary predictions. These studies further corroborated the importance of employing contour information to improve saliency predictions.

\begin{figure*}[!ht]
\begin{center}
\includegraphics[width=.99\linewidth]{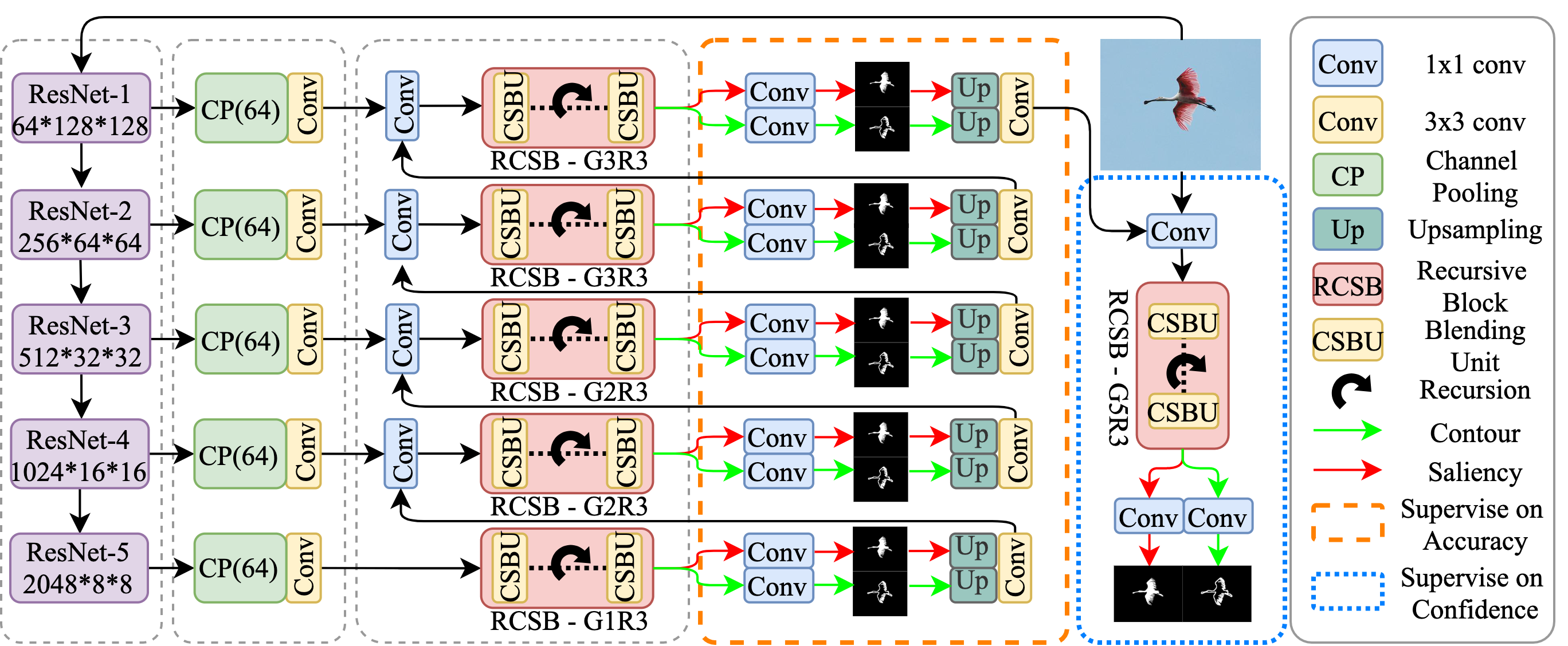}
\end{center}
\vspace{-0.02\linewidth}
\ \ \ \ \ Backbone\ \ \ \ \ \ \ \ \ \ 
\begin{minipage}{0.1\linewidth}
Channel Pooling
\end{minipage}
\begin{minipage}{0.8\linewidth}
\newcommand{\Repeat}[2]{%
    \foreach \n in {1,...,#1}{#2}
}
\Repeat{13}{\ }RCSB
\Repeat{34}{\ }SFE
\Repeat{23}{\ }Refinement
\Repeat{14}{\ }Legend
\end{minipage}

\caption{Network architecture for RCSBNet. Pre-trained ResNet-50 is used as the backbone, channel number is reduced to 64 via Channel Pooling (CP) layer. Recursive Contour-Saliency Blocks (RCSB) with $G$ blocks and $R$ recursions are then attached and followed by Stage-wise Feature Extraction (SFE) module to generate contour and saliency predictions. At the end of the network, a refinement module is adopted to further refine the predictions. }
\label{fig:RCSBNet}
\end{figure*}

\section{Proposed Method}
\subsection{Overall Architecture}
In previous contour-related networks \cite{POOLNET}\cite{EGNET}\cite{ITSD}, either the contour was supervised in a separate branch to guide the saliency prediction, or it was fused with saliency stage by stage to achieve better boundary predictions. Both approaches gave promising results, but there are two major disadvantages: 1) \textit{late fusion}: contours are fused with saliency at the end of each stage. 2) \textit{limited fusion}: the number of fusion is limited by the number of U-Net stages. For \textit{late fusion}, we designed a Contour-Saliency Blending Unit (CSBU) so that contour and saliency information can be exchanged at a much earlier stage,  while for \textit{limited fusion}, a recursive mechanism was adopted to circumvent this constraint.

As shown in Fig. \ref{fig:RCSBNet}, the proposed RCSBNet is essentially a U-Net, where we employ pre-trained ResNet-50 as our encoder and a customized decoder adopting recursive CNNs. Contour and saliency are blended in the recursion block by the Contour-Saliency Blending Unit (CSBU).  Then saliency and contour features are split and fed into the Stage-wise Feature Extraction (SFE) module for supervised learning. At the last stage of the decoder, we concatenate the prediction of contour and saliency with the input image, followed by an additional recursive block, to generate the final predictions.

\subsection{ResNet-50 as the Encoder}
We employ the pre-trained ResNet-50 as our encoder. Since it contains a large number of feature maps in high-level blocks,  same as \cite{ITSD}, we apply channel pooling (CP) to reduce the number of channels to 64, and the operation can be expressed as:
\begin{equation}
CP = collect_{j\in [0, m-1]}(max_{k\in[0,\frac{n}{m}-1]}X^{j\times \frac{n}{m}+k})
\end{equation}
where $j, k$ are integers, and we divide total $n$ channels into $m$ groups then apply max-pooling. A convolution layer is attached after the channel pooling layer to prepare the features for further processing by the decoder, as shown in Fig. \ref{fig:RCSBNet}.

\subsection{Contour-Saliency Blending Unit (CSBU)}

\begin{figure*}[!h]
 \includegraphics[width=\linewidth]{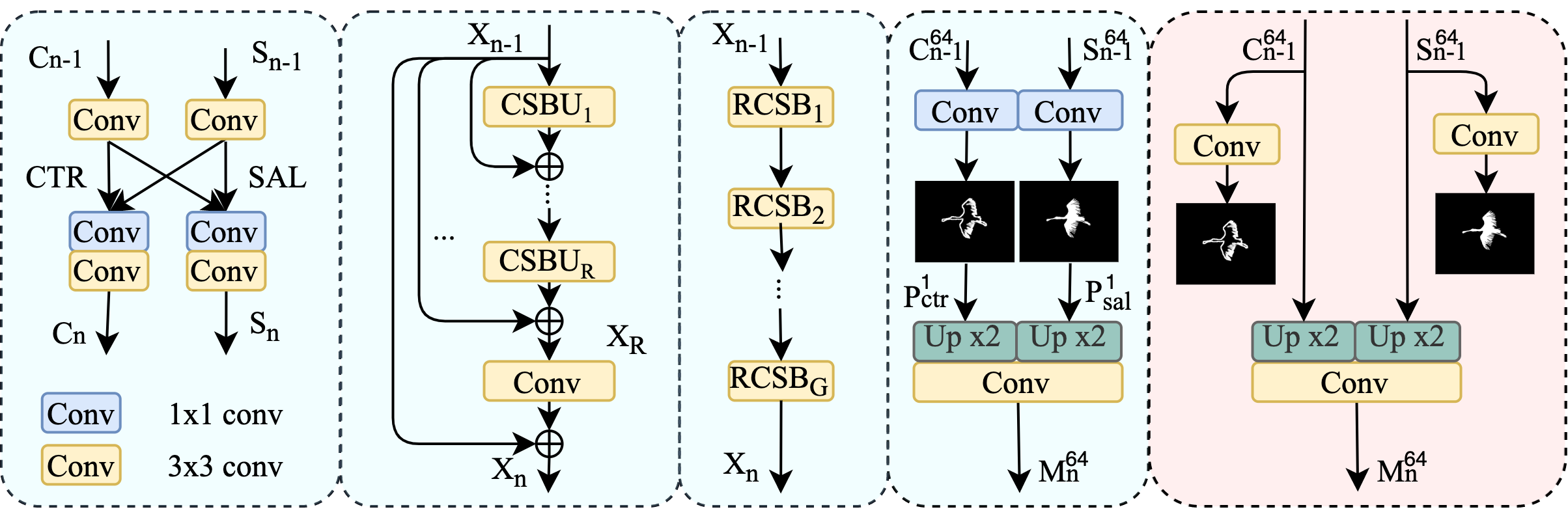}
\begin{subfigure}{.24\linewidth}
\caption{}
\label{fig:CSB}
\end{subfigure}
\begin{subfigure}{.18\linewidth}
\caption{}
\label{fig:CSBU}
\end{subfigure}
\begin{subfigure}{.13\linewidth}
\caption{}
\label{fig:RCSB}
\end{subfigure}
\begin{subfigure}{.16\linewidth}
\caption{}
\label{fig:SFE}
\end{subfigure}
\begin{subfigure}{.25\linewidth}
\caption{}
\label{fig:convention}
\end{subfigure}
\vspace{-0.02\linewidth}
\caption{(a) Contour-Saliency Blending Unit (CSBU). It contains two streams where contour and saliency information are blended and intertwined. (b) Single Recursive Contour-Saliency Blending (RCSB) Block. In order to increase the number of contour-saliency fusion, recursive mechanism is applied. Weights are shared among all CSBU blocks used in the RCSB. (c) Network branch contains $G$ blocks of RCSB, each with $R$ times of recursion. (d) Stage-wise Feature Extraction (SFE) module. (e) Conventional methods for generating intermediate stage predictions by using side branches. }
\label{fig:network_modules}
\end{figure*}

\label{CSB}
Contour information is usually supervised and fused with saliency at the end of each U-Net stage. For us, we want to fuse the contour and saliency at earlier stages and as many times as possible. Inspired by Shuffle-Net \cite{SHUFFLENET}, where convolution was divided into groups and information is exchanged by shuffling the weight channels, we designed our Contour-Saliency Blending Unit (CSBU). As illustrated in Fig. \ref{fig:CSB}, the CSBU has a contour branch and saliency branch. Features generated by each branch are concatenated and blended. Mathematically, let $F^{sal}_{k \times k}$ and $F^{ctr}_{k \times k}$ denote the convolution for saliency and contour with kernel size $k$ respectively, while ($C_{n-1}$, $S_{n-1}$) and ($C_{n}$, $S_{n}$) represent contour and saliency features before and after CSBU. For simplicity, we omit ReLU \cite{RELU} and BatchNorm \cite{BN} in our formula, and then the proposed CSBU can be modeled as:
\begin{equation}
CTR = F^{ctr}_{3 \times 3}(C_{n-1})
\end{equation}
\begin{equation}
SAL = F^{sal}_{3 \times 3}(S_{n-1})
\end{equation}
\begin{equation}
C_{n}= F^{ctr}_{3 \times 3}(F^{ctr}_{1 \times 1}([CTR,  SAL]))
\end{equation}
\begin{equation}
S_{n}= F^{sal}_{3 \times 3}(F^{sal}_{1 \times 1}([CTR,  SAL]))
\end{equation}
and thus:
\begin{equation}
S_{n}, C_{n}= CSBU(S_{n-1}, C_{n-1})
\label{eq_CSB}
\end{equation}

By applying CSBU, contour and saliency information can be utilized by the network at a much earlier stage.

\subsection{Recursive Block}
We now introduce more details of the recursive block, as illustrated in Fig. \ref{fig:CSBU}. Firstly let us consider a single recursive block then we extend to more general cases. We denote $R$ as the total number of recursions in a single block, and $f^r (r=1,2,..., R)$ denote the $r^{th}$ recursion of the CSBU. Based on Eq. \ref{eq_CSB}, let $X_{n-1}$ denote the input tuple of $(S_{n-1}, C_{n-1})$, and $X_{R}$ represent the output of the $R^{th}$ recursion, then we have:
\begin{equation}
X_{R} = f^{R}(...(f^{2}(f^{1}(X_{n-1}) + X_{n-1}) + X_{n-1}) + ... + X_{n-1})
\end{equation}
note that weights for CSBU are shared in recursion, though they have different superscripts in the formula. Then we apply a single convolution layer, denoted as $F_{3\times 3}$, at the end of recursion with skip connection. Thus the output of our recursive block, $X_{n}$, is:
\begin{equation}
X_{n} = F_{3\times 3}(X_{R} + X_{n-1}) + X_{n-1} = h_{b}(X_{n-1})
\end{equation}
where $h_b$ represents the recursive block function.

Now let $G$ represent the total number of recursive blocks,  as shown in Fig. \ref{fig:RCSB}, in our network we simply stack all the recursive blocks together. Thus the output of $g$-th block, $X_{n}$, is:
\begin{equation}
X_{n} = h^{g}_{b}(h^{g-1}_{b}(...h^2_{b}(h_{b}(X_{n-1}))))
\end{equation}
where $h^{g}_{b}$ represents the $g$-th recursive block function.

By applying recursive blocks and stacking them together, contour and saliency can now be fused $G\times R$ times at each stage of the U-Net, which will improve the network performance significantly. We will show more results in Sec. \ref{Experiments}.

\subsection{Stage-wise Feature Extraction (SFE) Module}

It is prevalent that saliency networks are densely supervised, where intermediate stage features are generated and supervised against ground truths to guide the model for better convergence. Common practices create a side branch with a few convolution layers to generate predictions from current U-Net stage features, following which the stage features are passed on to the next U-Net stage in the primary network (Fig. \ref{fig:convention}). Different from others, we regard stage predictions as the best result the network can obtain so far and apply a new round of feature extraction based on current stage predictions. Therefore, stage predictions are now in the network's main branch, acting as a single channel layer, and supervised against ground truths, as illustrated in Fig. \ref{fig:SFE}. By doing so, the next block is expected to extract valuable features from current best results and discard all useless features, which might otherwise be carried along by recursion and residual connections.

To generate stage predictions, unlike previous studies \cite{SHM}\cite{POOLNET}\cite{ITSD}, we do not use max-pooling because it will bring false positives to the next stage, nor the element-wise multiplication between contour and saliency like SCRN \cite{SCRN} due to the introduction of false negatives. Instead, we employ a $1\times 1$ convolution and a scaling factor to help sigmoid function classify saliency and background. To formulate the SFE module mathematically, let $C^{i}$ and $S^{i}$ represent incoming $i$ channels of feature maps, and $K$ represent the scaling factor learned by the network, then stage prediction $P^{j}$ with $j$ channels can be expressed as:
\begin{equation}
P^{1}_{sal} = F^{sal}_{1 \times 1}(S^{64}_{n-1}) * K_{sal}
\end{equation}
\begin{equation}
 P^{1}_{ctr} = F^{ctr}_{1 \times 1}(C^{64}_{n-1}) * K_{ctr}
\end{equation}
where * represents element-wise multiplication.  For extracted feature maps $M^{j}_{n}$ with $j$ channels:
\begin{equation}
M^{64}_{n} = F_{3 \times 3}(Up^{\times 2}[\sigma(P^{1}_{sal}), \sigma(P^{1}_{ctr})])
\end{equation}
where $[\ ]$ stands for concatenating, $Up^{\times 2}$ is up-sampling, and $\sigma$ stands for sigmoid function.
SFE is a simple module but it is very effective, we will domonstrate more in ablation studies.

\subsection{Loss Functions}

\begin{figure}[!h]
\centering
\begin{subfigure}{.32\linewidth}
\centering
\includegraphics[width=\linewidth]{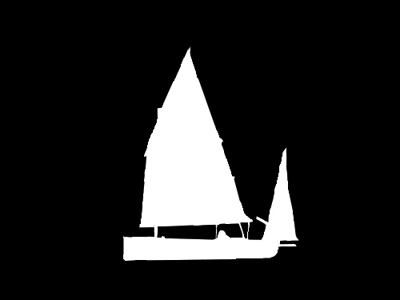}
\caption{}
\label{fig:gt}
\end{subfigure}
\begin{subfigure}{.32\linewidth}
\centering
\includegraphics[width=\linewidth]{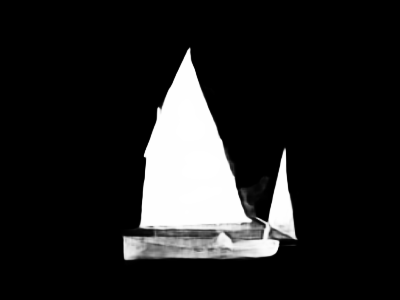}
\caption{}
\label{fig:pred}
\end{subfigure}
\begin{subfigure}{.32\linewidth}
\centering
\includegraphics[width=\linewidth]{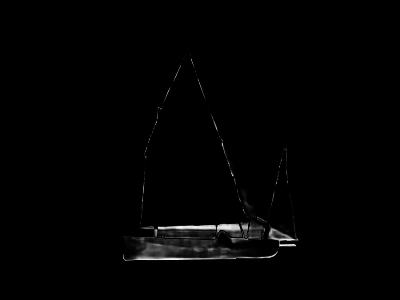}
\caption{}
\label{fig:focal_heatmap}
\end{subfigure}
\\
\begin{subfigure}{.32\linewidth}
\centering
\includegraphics[width=\linewidth]{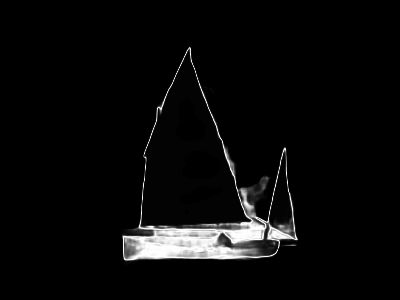}
\caption{}
\label{fig:confidence_heatmap}
\end{subfigure}
\begin{subfigure}{.32\linewidth}
\centering
\includegraphics[width=\linewidth]{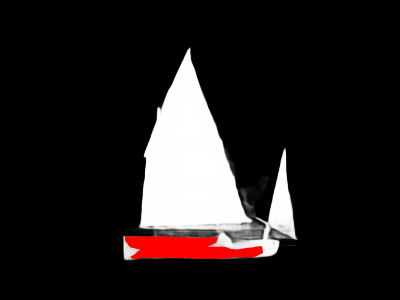}
\caption{}
\label{fig:unconfident}
\end{subfigure}
\begin{subfigure}{.32\linewidth}
\centering
\includegraphics[width=\linewidth]{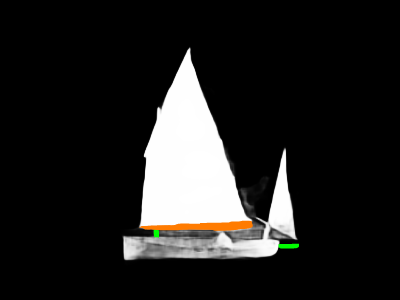}
\caption{}
\label{fig:confidentFPFN}
\end{subfigure}
\vspace{-0.01\linewidth}
\caption{(a) ground truth saliency. (b) predicted saliency. (c) focal loss ($\alpha, \gamma=2, 2$) weight map calculated using (a) and (b). (d) confidence loss ($\beta, \lambda=2, 1$) weight map calculated using (a) and (b). (e) unconfident but correct area (red) in prediction. (f) confident false positives (orange) and false negatives (green) in prediction. Compare (d) with (c), focal loss is less sensitive to unsure predictions,  while confidence loss generates high weights especially on object boundaries, which eventually help the model generate sharper edges.}
\label{fig:loss}
\end{figure}

Unsure prediction is very common in salient object detection in the form of shady areas (Fig. \ref{fig:pred}). Many studies try to improve the performance by focusing on hard pixels near the boundary \cite{CTLoss}\cite{F3NET}. Boundary pixels are indeed hard samples, but not all the hard samples are near the boundary. We notice that the network can correctly predict the saliency for some images, but with low pixel values; while for some other images, the network is confident of generating false negatives (FN) or false positives (FP), as illustrated in Fig. \ref{fig:unconfident} and \ref{fig:confidentFPFN}. It points out two kinds of difficulties encountered by the network: 1) unconfident but accurate predictions and 2) confident but inaccurate predictions. Hence in addition to accuracy,  we factor in the confidence of predictions, and we introduce Confidence Loss (CLoss) to our training.

\textbf{Confidence Loss}.
Recently focal loss \cite{FOCAL}\cite{focalSOD} has been explored in saliency tasks due to its high weight on wrong predictions: $W_{focal} = \alpha (\hat{y}_{i,j}-x_{i,j})^{\gamma}$. However, focal loss becomes less sensitive as predictions approach ground truth, which eventually leaves a large area of unsure predictions.

In order to guide the model focus more on the unconfident predictions, we propose a confidence score, $W_{c}$, for each predicted pixel $x_{i,j}$:
\begin{equation}
W_{c} = \beta * x_{i,j} * (1 - x_{i,j})
\end{equation}
where $\beta$ is empirically set to 2, and $x_{i,j}$ is the prediction after sigmoid. When $x_{i,j}=0.5$ the score reaches its maximum. Then with ground truth $\hat{y}_{i,j}$, our confidence loss (CLoss), $\mathcal{L}_{con}$, is defined as:
\begin{equation}
\mathcal{L}_{con}(\beta, \lambda) = -\frac{1}{n}\sum^{n}_{i=1}[W_{c} * BCE(x_{i,j}, \hat{y}_{i,j}) + \lambda W_{c}]
\end{equation}
where $\lambda$ is set to $0.3$ by parameter search. By applying this loss, it will encourage the model to make more confident predictions into either foreground (close to 1) or background (close to 0). 

\textbf{Dual Confinement Loss}.
Most of the contour-based models use a separate contour branch to guide saliency predictions. In \cite{CTLoss}, and \cite{ITSD}, a weight map generated from contour ground truth is applied to BCE loss to improve saliency boundary quality while the contour branch is supervised by BCE loss without any weights. It is reasonable because contour information is from a separate branch and is only used to guide saliency. Since in our network, saliency and contour streams are going to exchange information; thus we not only use contour to guide saliency predictions but also want to use saliency to guide contour predictions. Based on this, we designed our Dual Confinement Loss (DCLoss). The loss uses weight map generated by contour information to confine saliency predictions, and uses saliency information to generate weight map to guide contour predictions. This is due to the high correlation of contour and saliency information. The losses for each stream, $\mathcal{L}_{sal}$ and $\mathcal{L}_{ctr}$, are defined as:
\begin{equation}
\mathcal{L}_{sal} = -\frac{1}{n}\sum^{n}_{i=1}[W_{sal}*BCE(x^{sal}_{i}, \hat{y}^{sal}_{i})]
\end{equation}
\begin{equation}
 \mathcal{L}_{ctr} = -\frac{1}{n}\sum^{n}_{i=1}[W_{ctr}*BCE(x^{ctr}_{i}, \hat{y}^{ctr}_{i})]
\end{equation}
where weight matrix $W_{sal}$ and $W_{ctr}$ are calculated by:
\begin{equation}
W_{sal} = max(x^{ctr}_{i}, \hat{y}^{ctr}_{i}) * \theta + 1
\end{equation}
\begin{equation}
W_{ctr} = max(x^{sal}_{i}, \hat{y}^{sal}_{i}) * \theta + 1
\end{equation}
where $x, \hat{y}$ stand for prediction and ground truth, and empirically, we set $\theta=4$ in our experiments.
Then the DCLoss, $\mathcal{L}_{DC}$, is defined as:
\begin{equation}
\mathcal{L}_{DC} = \mathcal{L}_{sal} + \mathcal{L}_{ctr} 
\end{equation}
\textbf{Loss for Training}.
Compared with low-level predictions, it should be relatively easy to emphasize accuracy on high-level predictions due to its small feature dimension. Meanwhile, a failure in high-level predictions will impact its following decoders and eventually cause false positives or negatives. Thus, we train stages 1 to 5 of our network against accuracy-related losses, i.e. DCLoss and weighted IOU loss mentioned in \cite{F3NET}, and train the refinement module against CLoss only. Our loss functions for saliency and contour are defined as:
\begin{equation}
\mathcal{L}_{saliency} =  \mathcal{L}^{1-5}_{DC} + \mathcal{L}^{1-5}_{wIOU} + \mathcal{L}^{ref}_{con}(\beta, \lambda=2, 0.3)
\end{equation}
\begin{equation}
\mathcal{L}_{contour} =  \mathcal{L}^{1-5}_{DC} + \mathcal{L}^{ref}_{con}(\beta, \lambda=2, 0.3)
\end{equation}
where superscripts $1-5$ and $ref$ represent the five decoder stages and refinement module in Fig. \ref{fig:RCSBNet}.

\section{Experiments}
\label{Experiments}
\subsection{Datasets}
We use DUT-OMRON \cite{DUTO}(5168 images), ECSSD \cite{ECSSD}(1000 images), PASCAL-S \cite{PASCAL}(850 images), HKU-IS \cite{HKUIS}(4447 images), and DUTS-TE \cite{DUTSTE}(5019 Images) as our evaluation datasets.

\subsection{Implementation Details}
DUTS-TR \cite{DUTSTE} is used for training with input images resized to 256×256, then random horizontal flipping and $90^{\circ}$ rotation are applied as the augmentation. We use pre-trained ResNet-50 \cite{ResNet} as the encoder. The number of recursive blocks G is set to (1,2,2,3,3,5) with recursion R = 3 for the network (Fig. \ref{fig:RCSBNet}). Besides, we apply Leaky ReLU \cite{LEAKY} and adopt the Adam optimizer \cite{ADAM} with default hyperparameters to train our network. Learning rates for encoder and decoder are set to $10^{-5}$ and $10^{-4}$ respectively, and they are halved every 20 epochs with a total of 100 epochs using a batch size of 4. During testing, images are resized to 256×256, and the predictions (256×256) are resized back to their original size by using bilinear interpolation. We use PyTorch \cite{TORCH} and a single RTX 3090 GPU for our model and experiments. Code will be released soon.

\subsection{ Evaluation Metrics}
\label{eval_metric}
Precision-Recall (PR) curve, $F_{\beta}$-measure \cite{FBETA}, Mean Absolute Error (MAE), $F^{\omega}_{\beta}$-measure \cite{WFMEASURE}, and $E_{\xi}$-measure \cite{EMEASURE} are adopted in our experiments.

\textbf{PR-Curve}. By applying different thresholds from 0 to 255, PR curve is obtained by comparing the ground truth masks against the binarized saliency predictions. 

\textbf{F-measure}. The $F_{\beta}$-measure is calculated by precision and recall value of saliency maps:
$F_{\beta} = \frac{(1+\beta^2)\times Precision \times Recall}{\beta^2 \times Precision + Recall}$
where $\beta^2$ is set to 0.3 \cite{FBETA}. We report the average score over all thresholds from 0 to 255 and denote as $\overline{F_{\beta}}$ \cite{NLDF}\cite{BMPM}.

\textbf{MAE}. MAE is the mean value of the sum of pixel-wise absolute differences between predictions $x$ and ground truths $\hat{y}$:  $MAE = \frac{1}{n}\sum^{n}_{i=1}|x_i - \hat{y}_i|$.

\textbf{Weighted F-measure}. $F^{\omega}_{\beta}$ uses weighted precision and weighted recall to measure both exactness and completeness of the prediction against ground truth. It is designed to improve the existing $F_{\beta}$-measure.

\textbf{E-measure}. By using local pixel values and the image-wise mean, $E_{\xi}$ calculates the similarity between the prediction and the ground truth.

\subsection{Comparisons with State-of-the-art Results}
\label{compare}
We compare our results with 13 state-of-the-art salient object detection networks, including EGNet \cite{EGNET}, PoolNet \cite{POOLNET}, ITSD \cite{ITSD}, AFNet \cite{AFNET}, PAGE \cite{PAGE}, CPD \cite{CPD}, BASNet \cite{BASNET}, CAGNet \cite{CAGNET}, GateNet \cite{GATENET}, U2Net \cite{U2NET}, GCPA \cite{GCPA}, MINet \cite{MINET}, and F3Net \cite{F3NET}. Saliency maps used are provided by authors.

\begin{table*}[!h]
\scriptsize
\centering
\setlength\tabcolsep{3pt}
\caption{Quantitative comparisons between RCSBNet and other 13 methods on five benchmark datasets in terms of the average F-measure $\overline{F_{\beta}}$, MAE $M$, $E_{\xi}$ and $F^{\omega}_{\beta}$. $\uparrow$ / $\downarrow$ means the larger/smaller the value, the better the results. \textbf{\textcolor{red}{Red}}, \textbf{\textcolor{green}{Green}}, and \textbf{\textcolor{blue}{Blue}} indicate the best, second best and third best performance.}
\begin{tabular}{c|cccc|cccc|cccc|cccc|cccc} 
\hline
\hline
\multirow{2}{*}{\textbf{Method}} & \multicolumn{4}{c|}{\textbf{DUTS-TE}} & \multicolumn{4}{c|}{\textbf{HKU-IS}} & \multicolumn{4}{c|}{\textbf{PASCAL-S}} & \multicolumn{4}{c|}{\textbf{ECSSD}} & \multicolumn{4}{c}{\textbf{DUT-OMRON}}  \\
                        & \textbf{$\overline{F_{\beta}}\uparrow$} & \textbf{$M\downarrow$} & \textbf{$E_{\xi}\uparrow$} & \textbf{$F^{\omega}_{\beta}\uparrow$}             & \textbf{$\overline{F_{\beta}}\uparrow$} & \textbf{$M\downarrow$} & \textbf{$E_{\xi}\uparrow$}  & \textbf{$F^{\omega}_{\beta}\uparrow$}                 & \textbf{$\overline{F_{\beta}}\uparrow$} & \textbf{$M\downarrow$} & \textbf{$E_{\xi}\uparrow$}  & \textbf{$F^{\omega}_{\beta}\uparrow$}              & \textbf{$\overline{F_{\beta}}\uparrow$} & \textbf{$M\downarrow$} & \textbf{$E_{\xi}\uparrow$} & \textbf{$F^{\omega}_{\beta}\uparrow$}                & \textbf{$\overline{F_{\beta}}\uparrow$} & \textbf{$M\downarrow$} & \textbf{$E_{\xi}\uparrow$}  & \textbf{$F^{\omega}_{\beta}\uparrow$}              \\ 
\hline
\hline
\multicolumn{21}{c}{\textbf{Contour-based Methods}} \\
\hline
EGNet\textsubscript{19} \cite{EGNET} & \textbf{\textcolor{blue}{.815}}  & \textbf{\textcolor{blue}{.039}}  & .891   & .816 & \textbf{\textcolor{blue}{.901}}  & \textbf{\textcolor{blue}{.031}}  & .950  & .887 & \textbf{\textcolor{green}{.817}}   & .073   & \textbf{\textcolor{blue}{.848}}   & .795  & \textbf{\textcolor{green}{.920}}   & .037   & \textbf{\textcolor{green}{.927}}  & .903  & \textbf{\textcolor{blue}{.755}  } & \textbf{\textcolor{green}{.053}}   & \textbf{\textcolor{green}{.867}} & \textbf{\textcolor{blue}{.725}}  \\ 
\hline
PoolNet\textsubscript{19} \cite{POOLNET} & \textbf{\textcolor{green}{.819}}  & \textbf{\textcolor{green}{.037}}  & \textbf{\textcolor{green}{.896}}  & \textbf{\textcolor{blue}{.817}}  & \textbf{\textcolor{green}{.903}}  & \textbf{\textcolor{green}{.030}}  & \textbf{\textcolor{green}{.953}}  & \textbf{\textcolor{blue}{.889}}  & \textbf{\textcolor{green}{.826}}   & \textbf{\textcolor{green}{.064}}   & \textbf{\textcolor{red}{.852}}   & \textbf{\textcolor{blue}{.809}}  & \textbf{\textcolor{blue}{.919}}   & \textbf{\textcolor{blue}{.035}}   & \textbf{\textcolor{green}{.925}}  & \textbf{\textcolor{blue}{.904}} & .752  & \textbf{\textcolor{blue}{.054}}   & \textbf{\textcolor{red}{.868}}  & \textbf{\textcolor{blue}{.725}}  \\ 
\hline
ITSD\textsubscript{20} \cite{ITSD}    & .804   & .041   & \textbf{\textcolor{blue}{.895}}  & \textbf{\textcolor{green}{.824}} & .899  &  \textbf{\textcolor{blue}{.031}}   & \textbf{\textcolor{blue}{.952}}  & \textbf{\textcolor{green}{.894}}  & .785   & \textbf{\textcolor{blue}{.065}}   & \textbf{\textcolor{green}{.850}}  & \textbf{\textcolor{green}{.812}}  & .895 & \textbf{\textcolor{green}{.034}}   & \textbf{\textcolor{red}{.927}} & \textbf{\textcolor{green}{.911}}  & \textbf{\textcolor{green}{.756}}   & .061   & \textbf{\textcolor{blue}{.863}}   &  \textbf{\textcolor{green}{.750}} \\ 
\hline
Ours                           & \textbf{\textcolor{red}{.855}} & \textbf{\textcolor{red}{.034}}  & \textbf{\textcolor{red}{.903}  }& \textbf{\textcolor{red}{.840}}  & \textbf{\textcolor{red}{.923}} & \textbf{\textcolor{red}{.027}}  & \textbf{\textcolor{red}{.954}} & \textbf{\textcolor{red}{.909}} & \textbf{\textcolor{red}{.842}}& \textbf{\textcolor{red}{.058}} & \textbf{\textcolor{red}{.852}}& \textbf{\textcolor{red}{.816}} & \textbf{\textcolor{red}{.927}} & \textbf{\textcolor{red}{.033}}& \textbf{\textcolor{blue}{.923}} & \textbf{\textcolor{red}{.916}} & \textbf{\textcolor{red}{.773}}& \textbf{\textcolor{red}{.045}} &  .855 & \textbf{\textcolor{red}{.752}} \\
\hline
\hline
\multicolumn{21}{c}{\textbf{Non-Contour-based Methods}} \\
\hline
AFNet\textsubscript{19} \cite{AFNET}   & .793   & .046   & .879  & .785  & .889   & .036   & .942  & .872  & .828   & .078   & .846 & .804 & .908   & .042   & .918  & .886  & .739   & .057   & .853  & .717  \\ 
\hline
PAGE\textsubscript{19} \cite{PAGE}    & .777   & .051   & .854  & .769 & .884   & .037   & .940  & .868  & .817   & .078   & .835   & .792  & .906   & .042   & .920   & .886  & .736   & .066   & .853  & .722  \\ 
\hline
CPD\textsubscript{19} \cite{CPD}     & .805   & .043   & .886   & .795  & .891   & .034   & .944   &.876  & .831   & .072   & .849   & .803  & .917   & \textbf{\textcolor{blue}{.037}}  & \textbf{\textcolor{green}{.924}}  & .898 & .747   & .056   & \textbf{\textcolor{blue}{.866}}  & .719 \\ 
\hline
BASNet\textsubscript{19} \cite{BASNET}  & .791   & .047   & .884   &  .803  & \textbf{\textcolor{blue}{.898}}   & .032  & .946   & .890 & .781   & .076   & .847  & .800  & .879   & \textbf{\textcolor{blue}{.037}}  & .921   & .904 & .756   & .056   & \textbf{\textcolor{green}{.869}}   & \textbf{\textcolor{green}{.751}}  \\ 
\hline
CAGNet\textsubscript{20} \cite{CAGNET}     & \textbf{\textcolor{blue}{.837}}   & .040   & .897   & .817 & \textbf{\textcolor{green}{.909}}   & .030  & .945   & .893  & \textbf{\textcolor{blue}{.833}}   & .066   & \textbf{\textcolor{green}{.857}}  & .808  & .921   & \textbf{\textcolor{blue}{.037}}   & .916   & .902 & .752   & \textbf{\textcolor{blue}{.054}}   & .856  & .728  \\ 
\hline
GateNet\textsubscript{20} \cite{GATENET} & .806   & .040  & .889   & .809 & \textbf{\textcolor{blue}{.898}}   & .033   & \textbf{\textcolor{blue}{.949}}  &.879 & .819  & .068   & \textbf{\textcolor{blue}{.852}}   & .797 & .916   & .040   & \textbf{\textcolor{green}{.924}} & .894 & .746   & .055   & .862  & .729  \\ 
\hline
U2Net\textsubscript{20} \cite{U2NET}   & .792   & .045   & .886  & .804 & .896  & .031   & .948  & .889 & .770   & .076   & .841  & .792 & .892 & \textbf{\textcolor{red}{.033}} & \textbf{\textcolor{green}{.924}}  & .910 & \textbf{\textcolor{blue}{.761}}   & \textbf{\textcolor{blue}{.054}}  & \textbf{\textcolor{red}{.870}} & \textbf{\textcolor{green}{.751}} \\ 
\hline
GCPA\textsubscript{20} \cite{GCPA}     & .817   &  .038   & .891   & .821  & \textbf{\textcolor{blue}{ .898}}   & .031   & \textbf{\textcolor{blue}{.949}}  & .888 & .826   &  \textbf{\textcolor{green}{.061}}   & .847   & .808  &  .919   & \textbf{\textcolor{green}{.035}}   & .920  & .903 & .748   & .056   & .860   & .734  \\ 
\hline
MINet\textsubscript{20} \cite{MINET}     & .828   &  \textbf{\textcolor{blue}{.037}}   & \textbf{\textcolor{blue}{.898}}  & \textbf{\textcolor{blue}{.825}}  & \textbf{\textcolor{green}{.909}}   & \textbf{\textcolor{blue}{.029}}   & \textbf{\textcolor{green}{.953}}   &  \textbf{\textcolor{blue}{.897}} & .829   & .063   & .851  & \textbf{\textcolor{blue}{.809}}  & \textbf{\textcolor{blue}{.924}}   & \textbf{\textcolor{red}{.033}}   & \textbf{\textcolor{red}{.927}}   & \textbf{\textcolor{blue}{.911}} & .755   & .055   & .865   & .738 \\ 
\hline
F3Net\textsubscript{20} \cite{F3NET}   & \textbf{\textcolor{green}{.839}} & \textbf{\textcolor{green}{.035}} & \textbf{\textcolor{green}{.902}} & \textbf{\textcolor{green}{.835}} &  \textbf{\textcolor{green}{.909}} &  \textbf{\textcolor{green}{.028}} &  \textbf{\textcolor{green}{.953}}   & \textbf{\textcolor{green}{.900}} & \textbf{\textcolor{green}{.835}} & \textbf{\textcolor{blue}{.062}}   & \textbf{\textcolor{red}{.859}}   & \textbf{\textcolor{red}{.816}}  & \textbf{\textcolor{green}{.925}} & \textbf{\textcolor{red}{.033}} & \textbf{\textcolor{red}{.927}}  & \textbf{\textcolor{green}{.912}}   & \textbf{\textcolor{green}{.766}} & \textbf{\textcolor{green}{.053}}   & \textbf{\textcolor{red}{.870}} & \textbf{\textcolor{blue}{.747}} \\ 
\hline
Ours                           & \textbf{\textcolor{red}{.855}} & \textbf{\textcolor{red}{.034}}  & \textbf{\textcolor{red}{.903}} & \textbf{\textcolor{red}{.840}} & \textbf{\textcolor{red}{.923}} & \textbf{\textcolor{red}{.027}}  & \textbf{\textcolor{red}{.954}} & \textbf{\textcolor{red}{.909}} & \textbf{\textcolor{red}{.842}}& \textbf{\textcolor{red}{.058}} & \textbf{\textcolor{blue}{.852}}& \textbf{\textcolor{red}{.816}} & \textbf{\textcolor{red}{.927}} & \textbf{\textcolor{red}{.033}}& \textbf{\textcolor{blue}{.923}} & \textbf{\textcolor{red}{.916}} & \textbf{\textcolor{red}{.773}}& \textbf{\textcolor{red}{.045}} &  .855& \textbf{\textcolor{red}{.752}} \\
\hline
\hline
\end{tabular}
\label{table:MAE}
\end{table*}


\begin{figure*}[!h]
\centering
\begin{subfigure}{\linewidth}
\centering
\newcommand\id{1189}
\includegraphics[width=.07\linewidth]{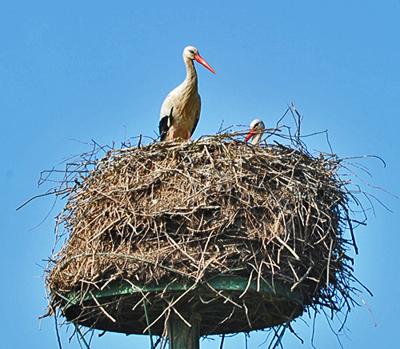}  
\includegraphics[width=.07\linewidth]{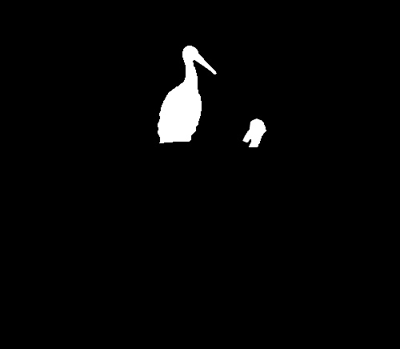}  
\includegraphics[width=.07\linewidth]{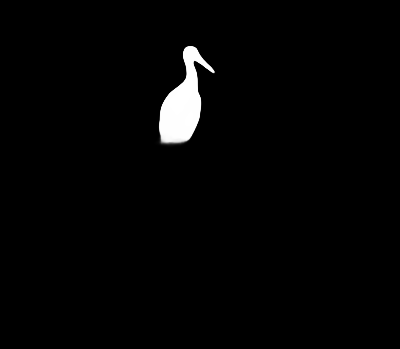}  
\includegraphics[width=.07\linewidth]{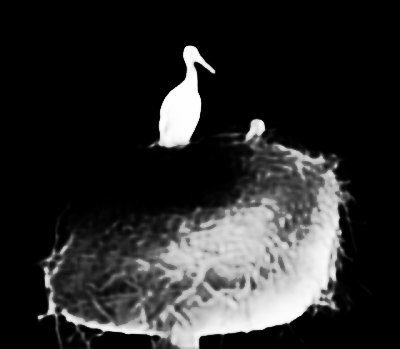}  
\includegraphics[width=.07\linewidth]{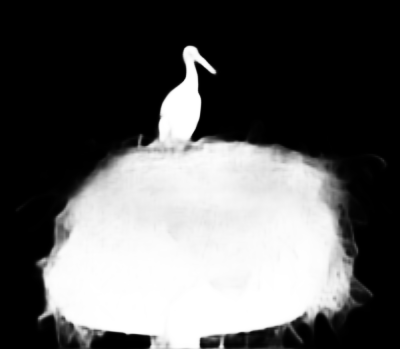}  
\includegraphics[width=.07\linewidth]{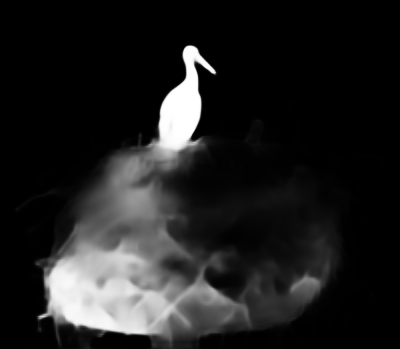}  
\includegraphics[width=.07\linewidth]{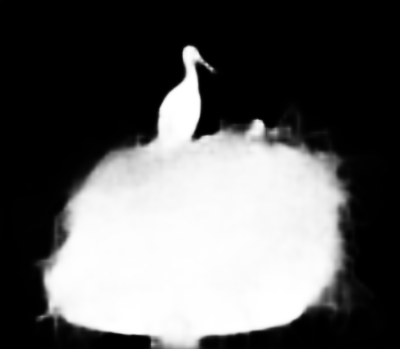}  
\includegraphics[width=.07\linewidth]{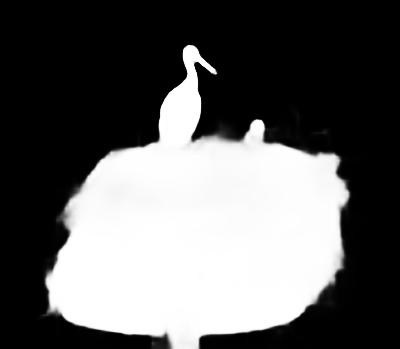}  
\includegraphics[width=.07\linewidth]{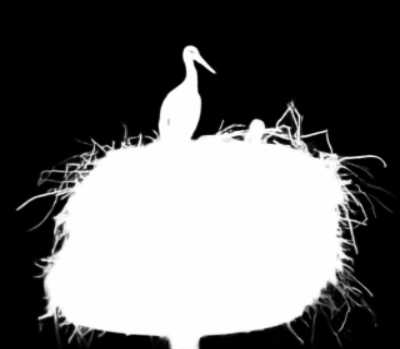}  
\includegraphics[width=.07\linewidth]{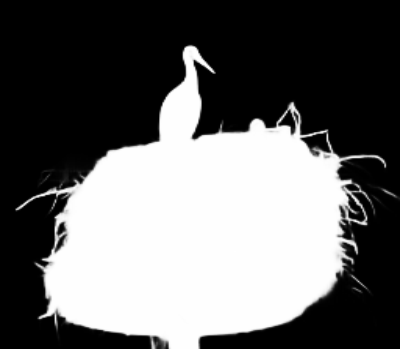}  
\includegraphics[width=.07\linewidth]{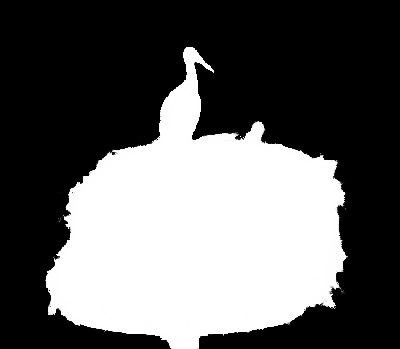}  
\includegraphics[width=.07\linewidth]{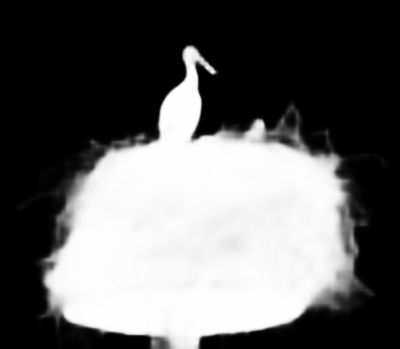}  
\includegraphics[width=.07\linewidth]{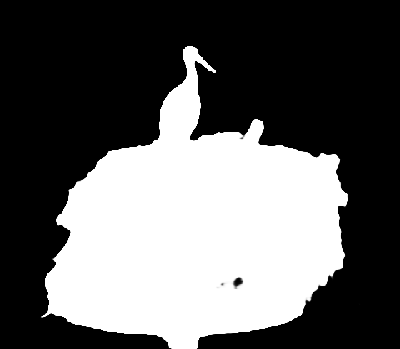}  
\end{subfigure}
\begin{subfigure}{\linewidth}
\centering
\newcommand\id{1354}
\includegraphics[width=.07\linewidth]{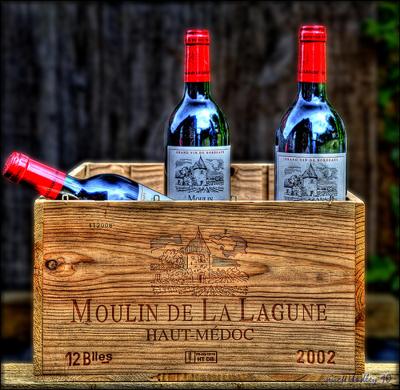}  
\includegraphics[width=.07\linewidth]{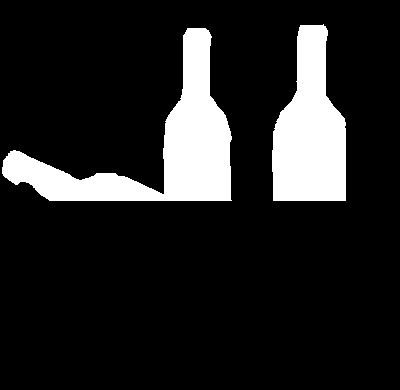}  
\includegraphics[width=.07\linewidth]{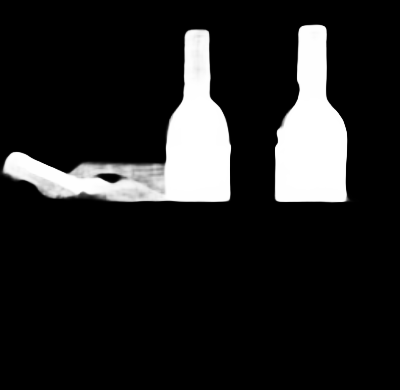}  
\includegraphics[width=.07\linewidth]{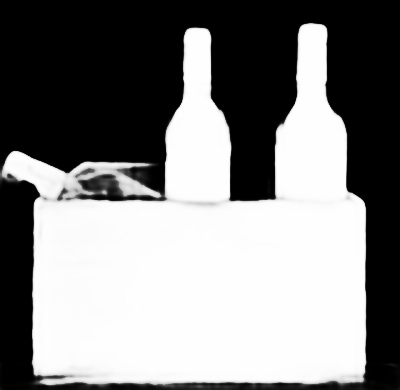}  
\includegraphics[width=.07\linewidth]{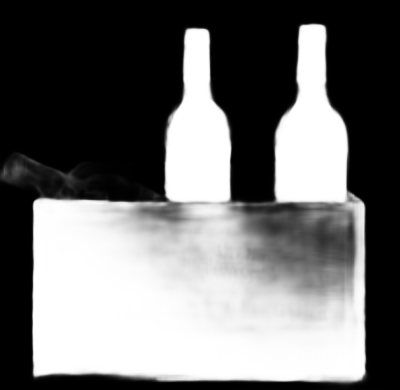}  
\includegraphics[width=.07\linewidth]{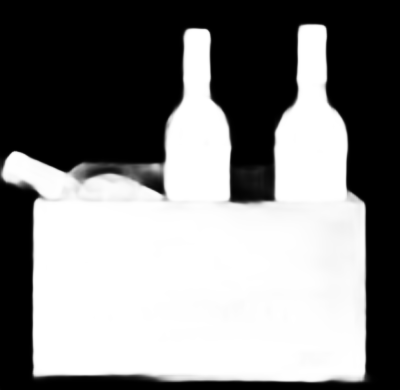}  
\includegraphics[width=.07\linewidth]{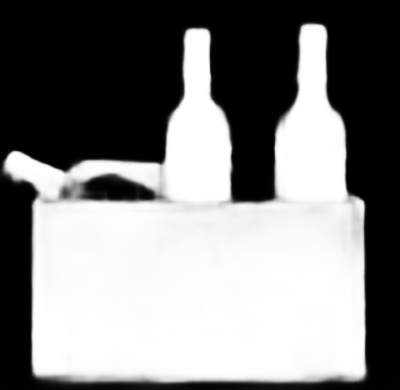}  
\includegraphics[width=.07\linewidth]{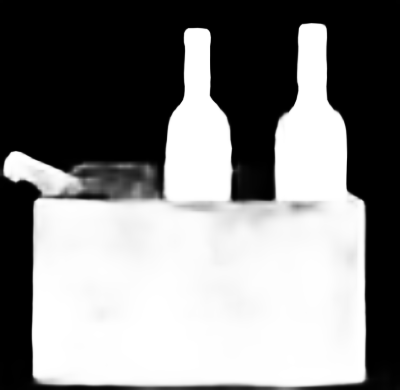}  
\includegraphics[width=.07\linewidth]{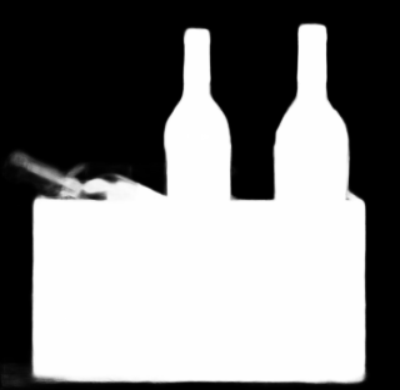}  
\includegraphics[width=.07\linewidth]{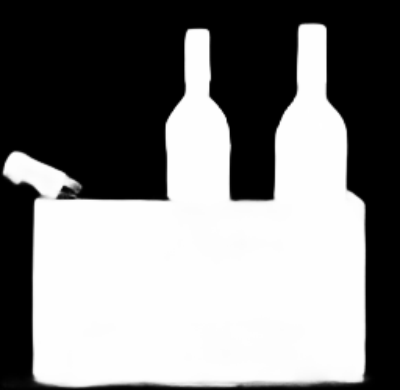}  
\includegraphics[width=.07\linewidth]{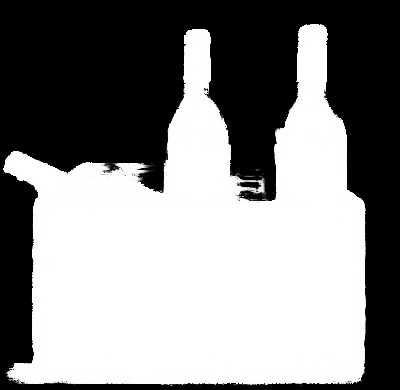}  
\includegraphics[width=.07\linewidth]{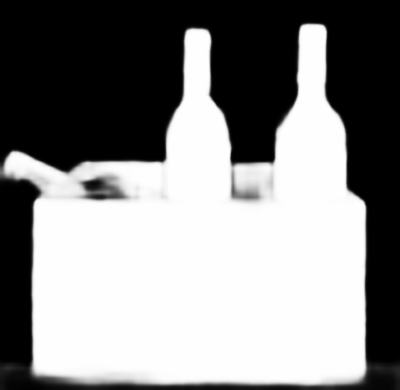}  
\includegraphics[width=.07\linewidth]{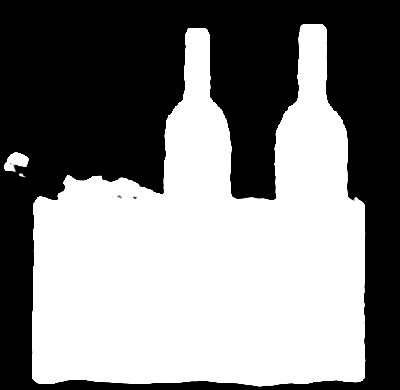}  
\end{subfigure}
\begin{subfigure}{\linewidth}
\centering
\newcommand\id{3220}
\includegraphics[width=.07\linewidth]{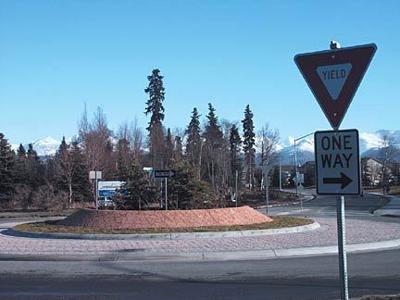}  
\includegraphics[width=.07\linewidth]{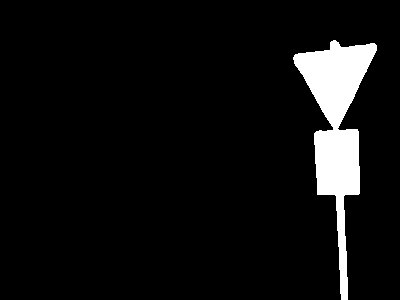}  
\includegraphics[width=.07\linewidth]{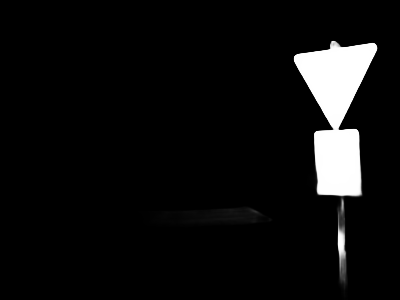}  
\includegraphics[width=.07\linewidth]{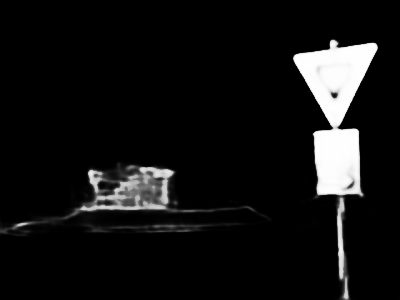}  
\includegraphics[width=.07\linewidth]{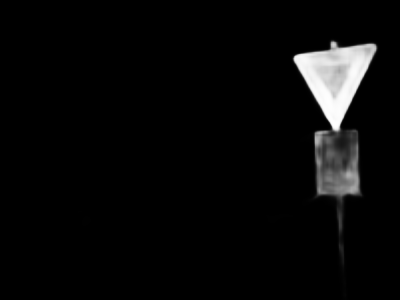}  
\includegraphics[width=.07\linewidth]{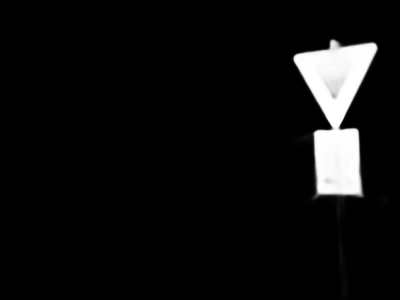}  
\includegraphics[width=.07\linewidth]{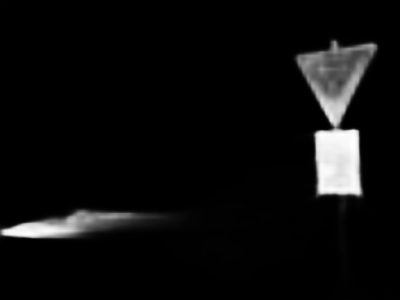}  
\includegraphics[width=.07\linewidth]{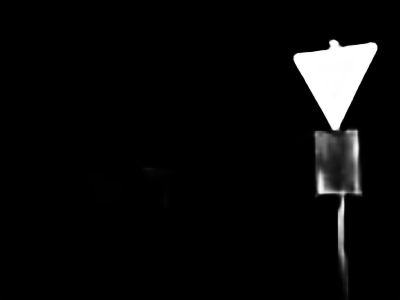}  
\includegraphics[width=.07\linewidth]{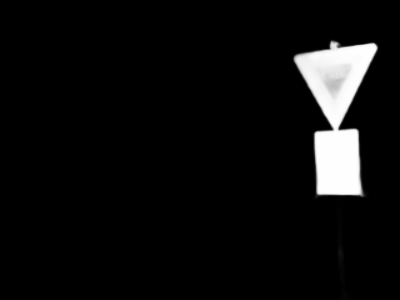}  
\includegraphics[width=.07\linewidth]{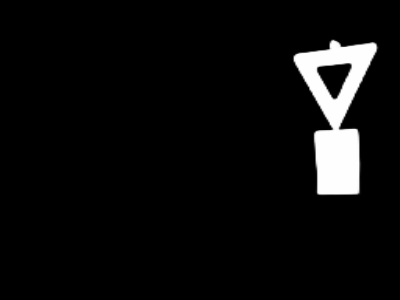}  
\includegraphics[width=.07\linewidth]{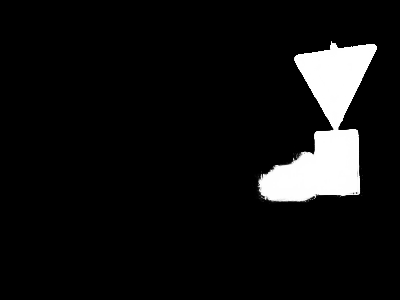}  
\includegraphics[width=.07\linewidth]{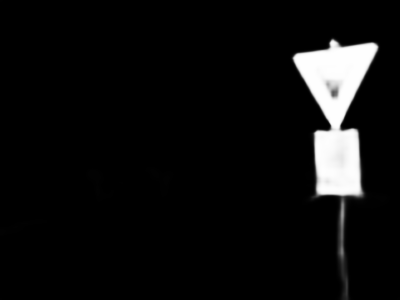}  
\includegraphics[width=.07\linewidth]{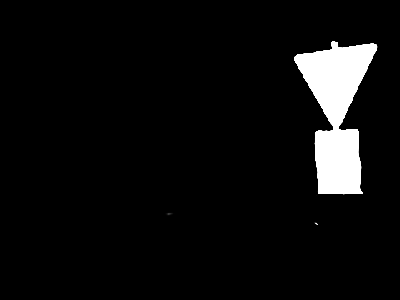}  
\end{subfigure}
\begin{subfigure}{\linewidth}
\centering
\newcommand\id{2777}
\includegraphics[width=.07\linewidth]{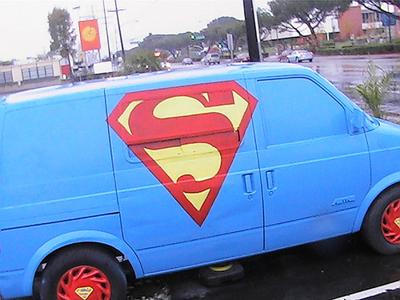}  
\includegraphics[width=.07\linewidth]{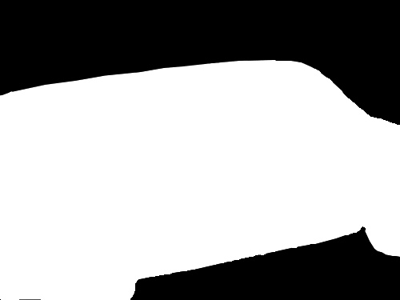}  
\includegraphics[width=.07\linewidth]{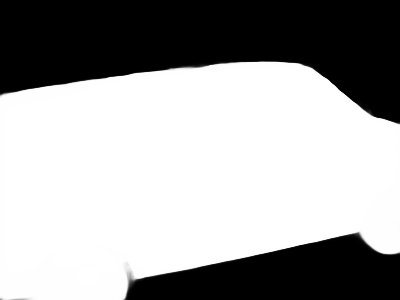}  
\includegraphics[width=.07\linewidth]{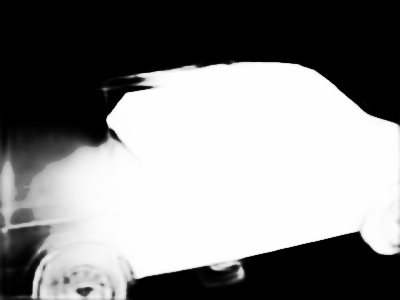}  
\includegraphics[width=.07\linewidth]{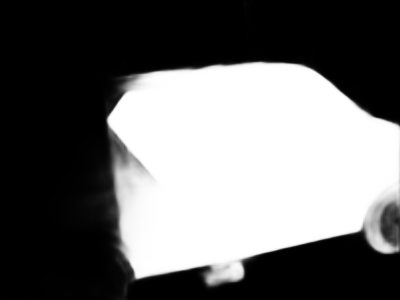}  
\includegraphics[width=.07\linewidth]{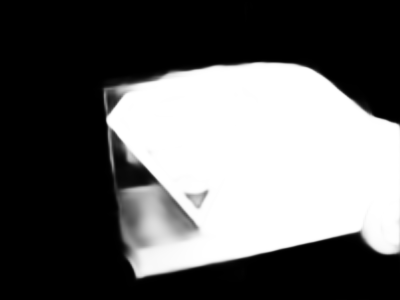}  
\includegraphics[width=.07\linewidth]{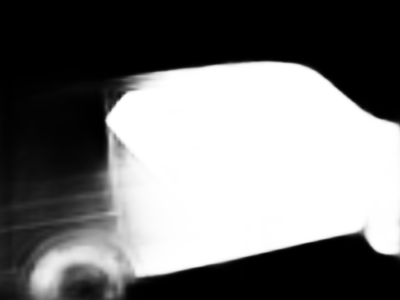}  
\includegraphics[width=.07\linewidth]{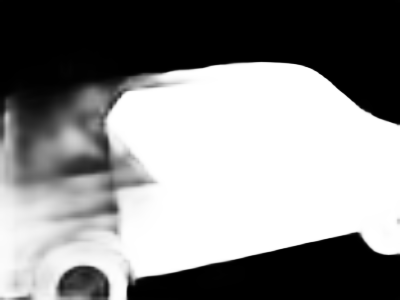}  
\includegraphics[width=.07\linewidth]{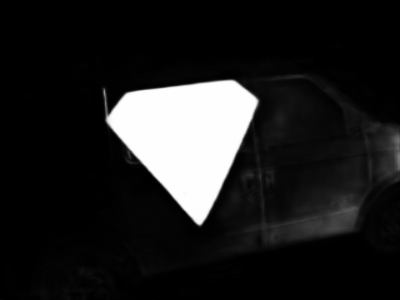}  
\includegraphics[width=.07\linewidth]{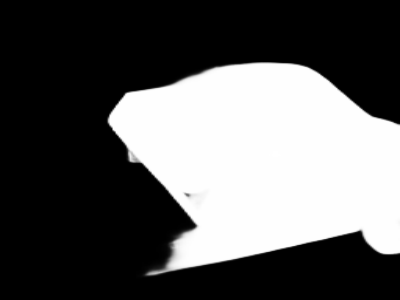}  
\includegraphics[width=.07\linewidth]{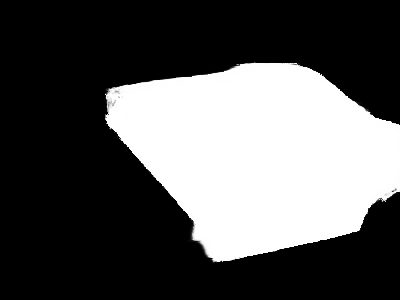}  
\includegraphics[width=.07\linewidth]{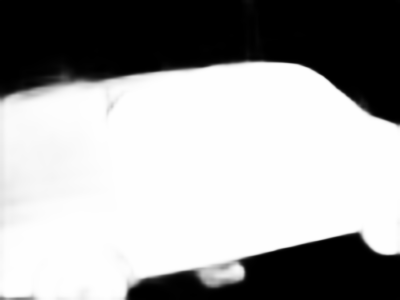}  
\includegraphics[width=.07\linewidth]{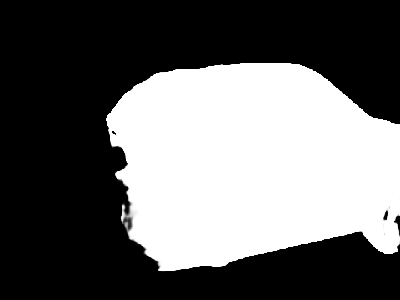}  
\end{subfigure}
\begin{subfigure}{\linewidth}
\centering
\newcommand\id{1106}
\includegraphics[width=.07\linewidth]{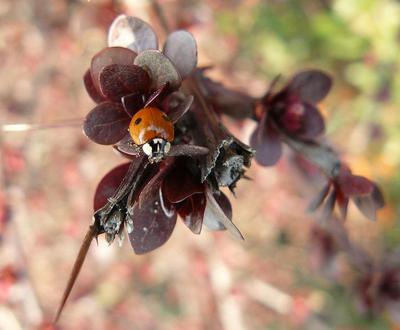}  
\includegraphics[width=.07\linewidth]{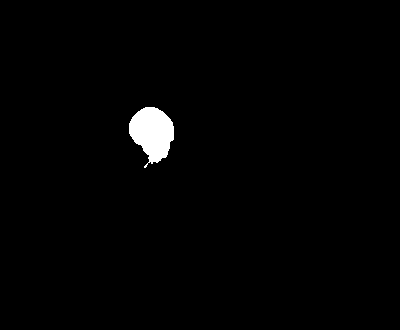}  
\includegraphics[width=.07\linewidth]{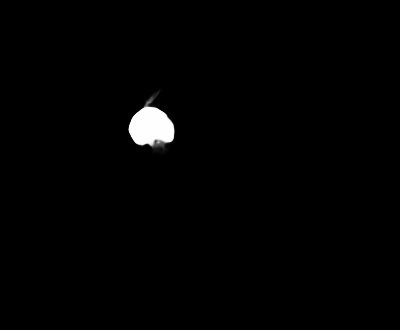}  
\includegraphics[width=.07\linewidth]{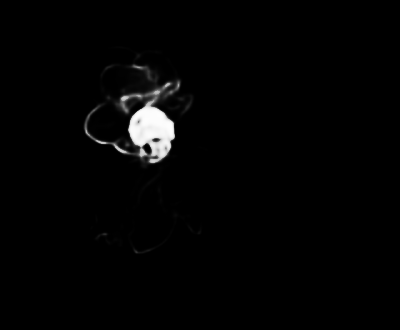}  
\includegraphics[width=.07\linewidth]{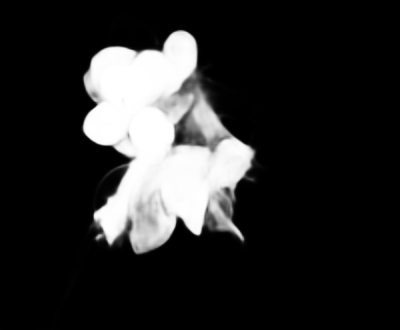}  
\includegraphics[width=.07\linewidth]{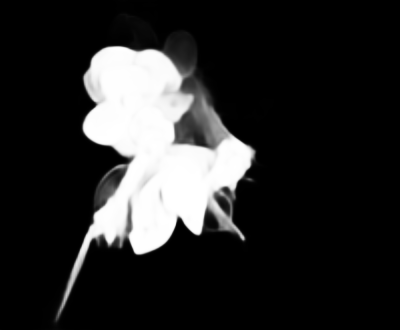}  
\includegraphics[width=.07\linewidth]{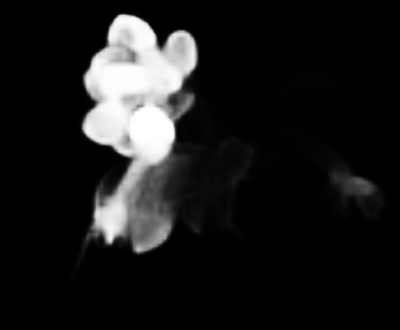}  
\includegraphics[width=.07\linewidth]{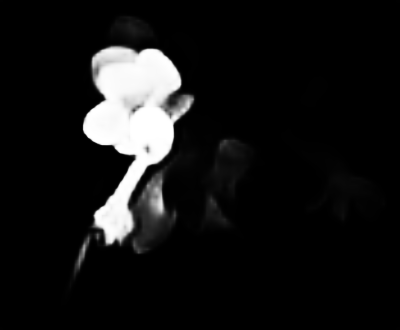}  
\includegraphics[width=.07\linewidth]{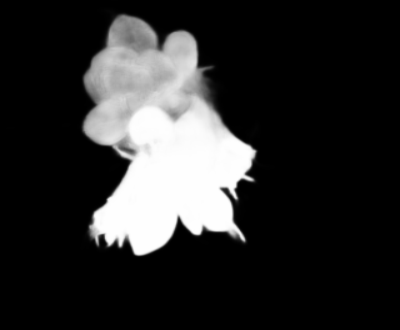}  
\includegraphics[width=.07\linewidth]{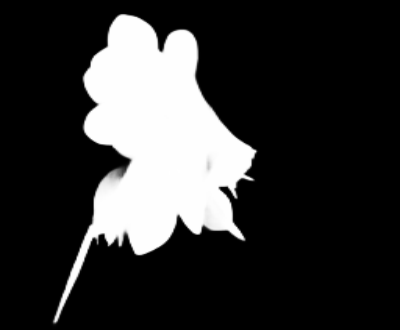}  
\includegraphics[width=.07\linewidth]{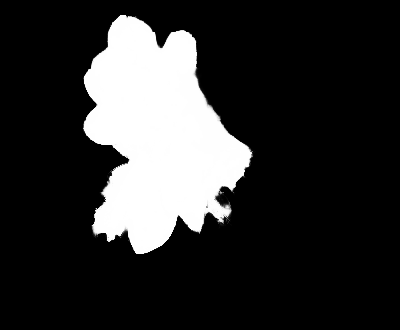}  
\includegraphics[width=.07\linewidth]{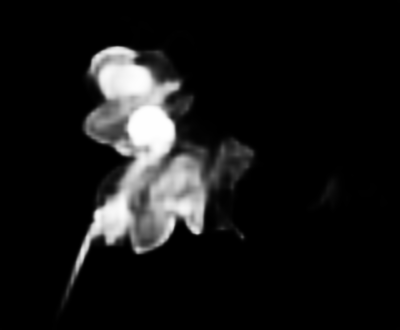}  
\includegraphics[width=.07\linewidth]{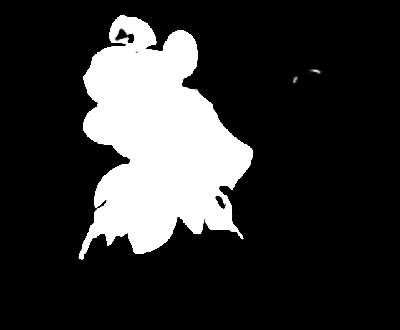}  
\end{subfigure}
\begin{minipage}{\linewidth}
{\footnotesize
\ \ \ \ \ \ \ \ Image
\ \ \ \ \ \ \ \ \ GT
\ \ \ \ \ \ \ \ \ \ \ Ours*
\ \ \ \ \ \ \ ITSD*
\ \ \ \ PoolNet*
\ \ \ \ \ EGNet*
\ \ \ \ \ MINet
\ \ \ \ \ \ F3Net
\ \ \ \ \ \ \ U2Net
\ \ \ \ \ BASNet
\ \ \ \ \ \ PAGE
\ \ \ \ \ \ \ \ GCPA
\ \ \ \ \ CAGNet
}
\end{minipage}
\caption{Visual comparisons between our method and 10 state-of-the-art networks. * stands for models utilizing contour information. More comparisons are provided in the supplementary material.}
\label{fig:visualization}
\end{figure*}

\begin{figure*}[!h]
\centering
     \includegraphics[width=\linewidth]{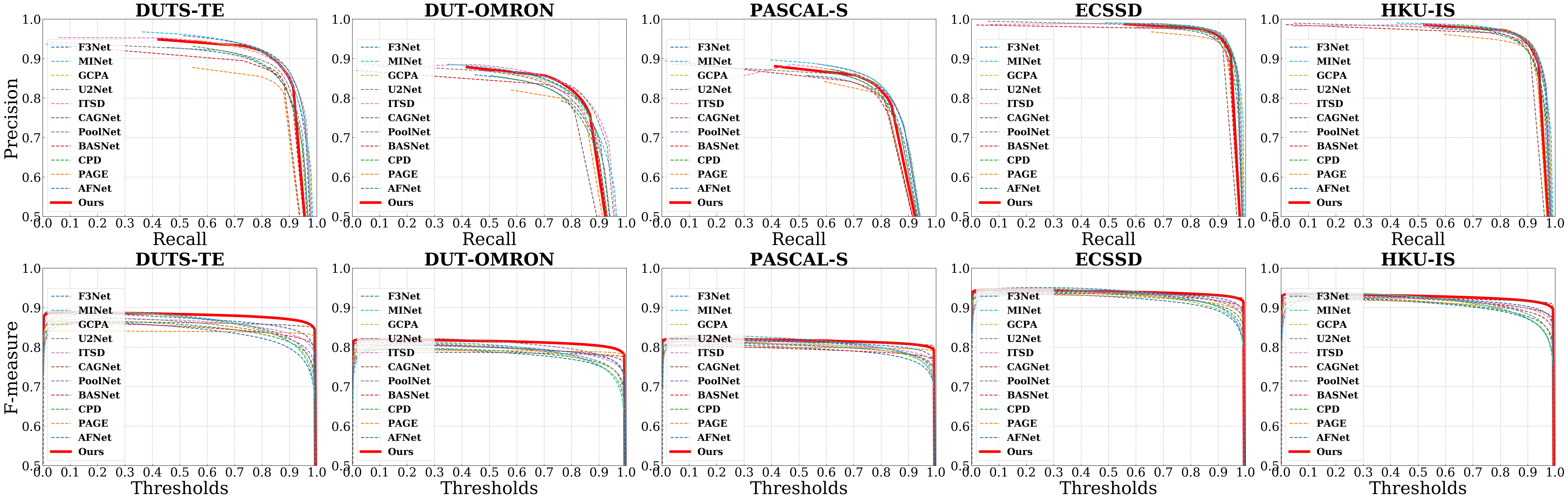}
     \caption{First row: Precision-Recall Curves comparison on five saliency benchmark datasets. Second row: F-measure Curves comparison on five saliency benchmark datasets. }
     \label{fig:PR}
\end{figure*}

\textbf{Quantitative Evaluation.}
To compare our work with the state-of-the-art networks, detailed experimental results in terms of four metrics are listed in Table \ref{table:MAE}. Among all the models, RCSBNet achieves outstanding results across all four metrics on most datasets. Besides, PR and F-measure curves are demonstrated in Fig. \ref{fig:PR}. Our F-measure curves are flatter than all other models, which reveals that our results are closer to binary predictions and invariant to threshold changes.

\textbf{Qualitative Evaluation.}
Visual comparisons are listed in Fig. \ref{fig:visualization}. Compare with other contour-based network results, our method yields better boundary predictions. As shown in the graph, our model can produce accurate and complete saliency maps with better edges.


\subsection{Ablation Studies}
\label{ablation}
\begin{figure}[!h]
\centering
\begin{subfigure}{.19\linewidth}
\centering
\includegraphics[width=\linewidth]{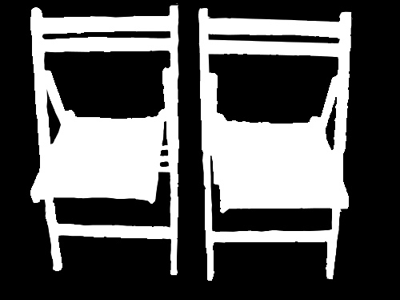} 
\includegraphics[width=\linewidth]{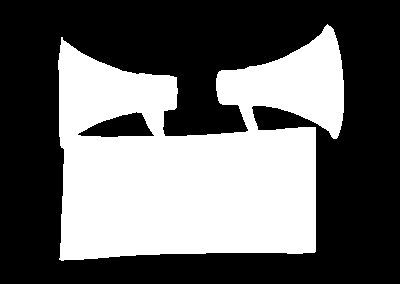}  
\caption{GT}
\label{fig:wef_sal}
\end{subfigure}
\begin{subfigure}{.19\linewidth}
\centering
\includegraphics[width=\linewidth]{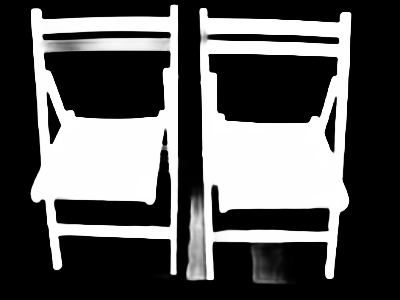} 
\includegraphics[width=\linewidth]{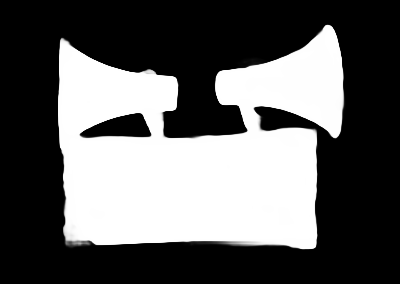}  
\caption{w/ EF}
\label{fig:wef_sal}
\end{subfigure}
\begin{subfigure}{.19\linewidth}
\centering
\includegraphics[width=\linewidth]{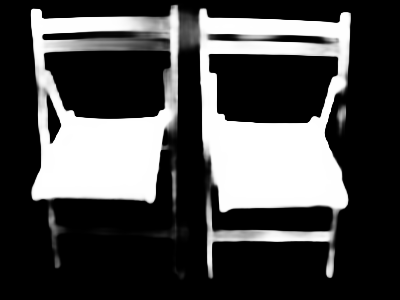} 
\includegraphics[width=\linewidth]{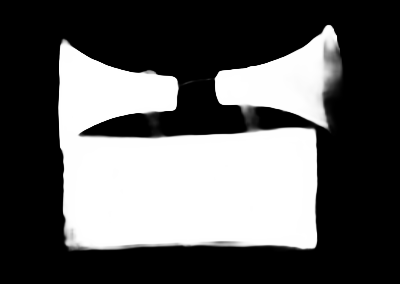}  
\caption{w/o EF}
\label{fig:woef_sal}
\end{subfigure}
\begin{subfigure}{.19\linewidth}
\centering
\includegraphics[width=\linewidth]{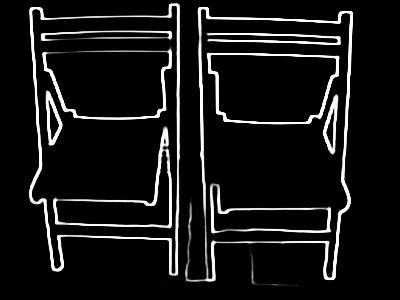} 
\includegraphics[width=\linewidth]{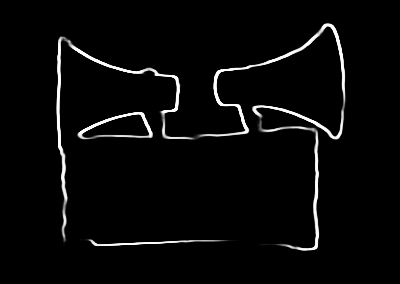}  
\caption{w/ EF}
\label{fig:wef_ctr}
\end{subfigure}
\begin{subfigure}{.19\linewidth}
\centering
\includegraphics[width=\linewidth]{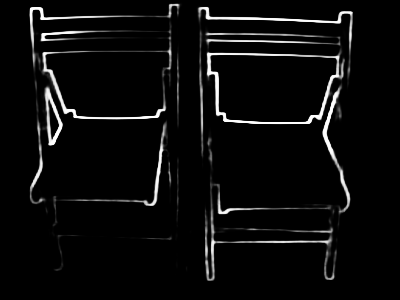}  
\includegraphics[width=\linewidth]{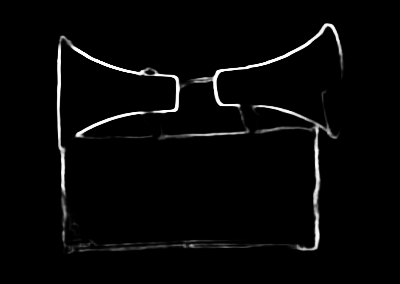}
\caption{w/o EF}
\label{fig:woef_ctr}
\end{subfigure}
\vspace{-0.03\linewidth}
\caption{Qualitative comparisons for early fusion.  (a) Ground-truth. (b) \& (d) Saliency and contour prediction with early fusion. (c) \& (e) Saliency and contour prediction without early fusion.  As can be seen with early fusion, results are more complete for both contour and saliency predictions.}
  \label{fig:early_fusion}
\end{figure}

\textbf{Effectiveness of the early fusion (EF)}. To prove that early fusion of contour and saliency information will boost model performance, an experiment was conducted by removing the fusion branch illustrated in Fig. \ref{fig:CSB} and make it into two seperate streams. We list both quantitative and qualitative measures between the two approaches on DUTS-TE and ECSSD datasets, as shown in Table \ref{table:early_fusion} and Fig. \ref{fig:early_fusion}.

\vspace{-0.05\linewidth}

\begin{table}[!h]
      \centering
	\setlength\tabcolsep{0pt}
	\caption{Quantitative comparisons for the effectiveness of early fusion.}
        \begin{tabular}{c|cccc|cccc} 
		\hline
		\multirow{2}{*}{} & \multicolumn{4}{c|}{DUTS-TE} & \multicolumn{4}{c}{ECSSD}  \\
		                           & \textbf{$\overline{F_{\beta}}\uparrow$} & \textbf{$M\downarrow$} & \textbf{$E_{\xi}\uparrow$} & \textbf{$F^{\omega}_{\beta}\uparrow$}                & \textbf{$\overline{F_{\beta}}\uparrow$} & \textbf{$M\downarrow$} & \textbf{$E_{\xi}\uparrow$} & \textbf{$F^{\omega}_{\beta}\uparrow$} \\ 
		\hline
		w/ early fusion        &\textbf{.855}   & \textbf{.034}  & \textbf{.903}  & \textbf{.840} & \textbf{.927} & \textbf{.033}  & \textbf{.923}  & \textbf{.916}    \\
		w/o early fusion      & .849               & .037                 & .897                & .833               & .925               & .035               & .919                & .908          \\
		\hline
	\end{tabular}
	\label{table:early_fusion}
\end{table}

\textbf{Effectiveness of refinement module and supervised on confidence}. To study the importance of the refinement module and prove the effectiveness of supervision on confidence, we conducted 4 experiments on DUTS-TE and ECSSD datasets covering all the cases for our comparison, as listed in Table \ref{table:ref_and_conf}. For conciseness, we denote reference module and supervision on confidence as \textit{Ref.} and \textit{Conf.}.

\begin{table}[!h]
      \centering
	\setlength\tabcolsep{1pt}
	\caption{Quantitative comparisons for different model configurations.}
        \begin{tabular}{c|c|cccc|cccc} 
		\hline
		\multirow{2}{*}{Ref.} & \multirow{2}{*}{Conf.}  & \multicolumn{4}{c|}{DUTS-TE} & \multicolumn{4}{c}{ECSSD}  \\
		                                  &                                        & \textbf{$\overline{F_{\beta}}\uparrow$} & \textbf{$M\downarrow$} & \textbf{$E_{\xi}\uparrow$} & \textbf{$F^{\omega}_{\beta}\uparrow$}                & \textbf{$\overline{F_{\beta}}\uparrow$} & \textbf{$M\downarrow$} & \textbf{$E_{\xi}\uparrow$} & \textbf{$F^{\omega}_{\beta}\uparrow$} \\ 
		\hline
		\xmark & \xmark   & .842   & .039  & .849  & .825  &  .916 &  .037   &  .914  &  .902    \\
		\cmark &\xmark    & .848   &  .038  & .861  &  .830 &  .920 &  .036   &  .916  &  .908    \\
		\xmark &\cmark    & .849   &  .037  & .870  &  .837 &  .922 &  .035   &  .921  &  .909    \\
		\cmark  &  \cmark &\textbf{.855}   & \textbf{.034}  & \textbf{.903}  & \textbf{.840} & \textbf{.927} & \textbf{.033}  & \textbf{.923}  & \textbf{.916}         \\
		\hline
	\end{tabular}
	\label{table:ref_and_conf}
\end{table}

It can be observed that each approach will boost the performance, and when they are combined, we obtained the best results.

\textbf{Effectiveness of SFE module and different loss functions}. To study the importance of each loss function and SFE module, we conduct a series of controlled experiments on the DUTS-TE dataset. We train the model by using BCE loss only, then include weighted IOU Loss \cite{F3NET}, DCLoss, and CLoss step by step. Detailed results are listed in Table \ref{table:loss_and_sfe}.

\begin{table}[!h]
      \centering
        \setlength\tabcolsep{1.5pt}
	\caption{Ablation study for different loss functions and presence of SFE module.}
        \begin{tabular}{c|c|c|c|c|cccc}
	\hline
	\multirow{2}{*}{BCE} & \multirow{2}{*}{wIOU} & \multirow{2}{*}{DCLoss} & \multirow{2}{*}{CLoss} & \multirow{2}{*}{SFE} & \multicolumn{4}{c}{DUTS-TE}  \\
	                     &                        &                         &                        &                      & \textbf{$\overline{F_{\beta}}\uparrow$} & \textbf{$M\downarrow$} & \textbf{$E_{\xi}\uparrow$} & \textbf{$F^{\omega}_{\beta}\uparrow$} \\ 
	\hline
	\cmark &                        &                         &                        &                      &.788              & .058              & .862               & .776 \\
	\cmark & \cmark   &                         &                        &                      &.793              & .047               & .881               & .789 \\
	                     & \cmark   & \cmark    &                        &                      &.829              & .043              & .890               & .813 \\
	                     & \cmark   & \cmark    & \cmark   &                      &.847              & .040               & .896               & .825 \\
	                     & \cmark   & \cmark    & \cmark   & \cmark &\textbf{.855} & \textbf{.034}  & \textbf{.903}  & \textbf{.840} \\
	\hline
	\end{tabular}
	\label{table:loss_and_sfe}
\end{table}

\begin{figure}[!h]
\centering
\begin{subfigure}{.19\linewidth}
\centering
\includegraphics[width=\linewidth]{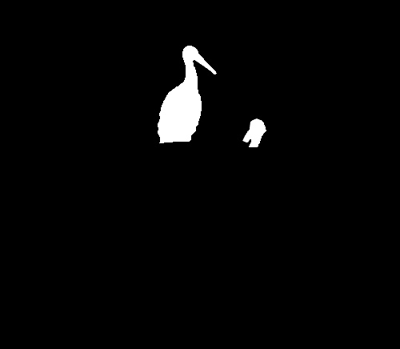} 
\includegraphics[width=\linewidth]{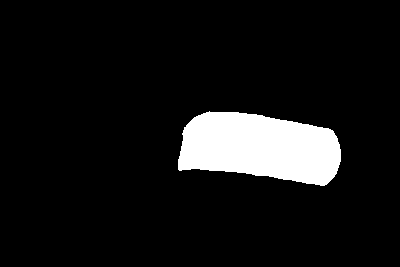}  
\caption{GT}
\label{fig:wef_sal}
\end{subfigure}
\begin{subfigure}{.19\linewidth}
\centering
\includegraphics[width=\linewidth]{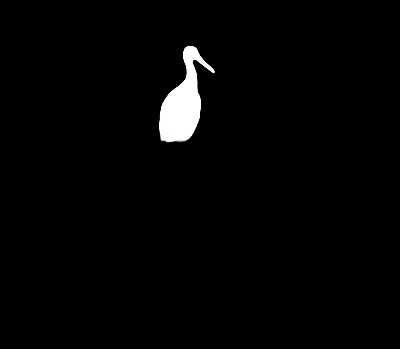} 
\includegraphics[width=\linewidth]{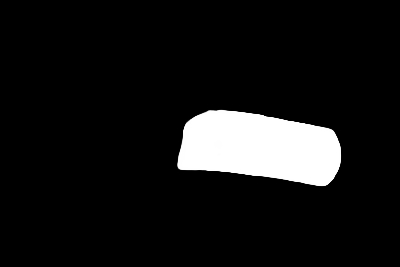}  
\caption{w/ SFE}
\label{fig:wef_sal}
\end{subfigure}
\begin{subfigure}{.19\linewidth}
\centering
\includegraphics[width=\linewidth]{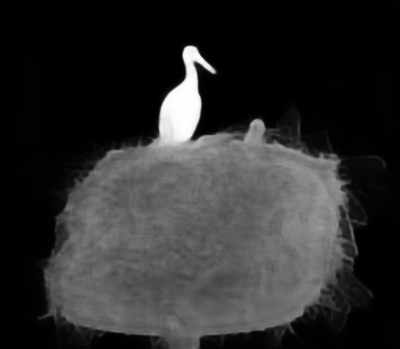} 
\includegraphics[width=\linewidth]{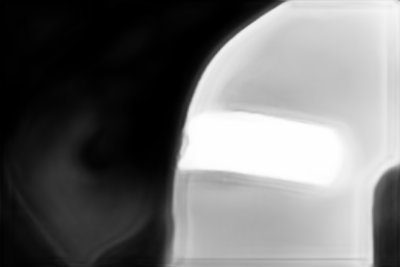} 
\caption{w/o SFE}
\label{fig:woef_sal}
\end{subfigure}
\begin{subfigure}{.19\linewidth}
\centering
\includegraphics[width=\linewidth]{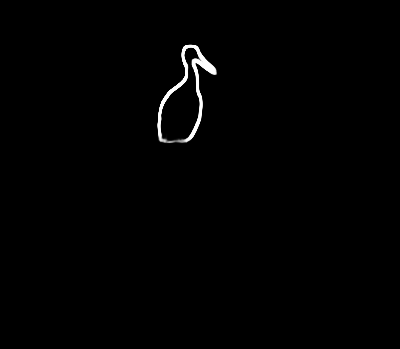} 
\includegraphics[width=\linewidth]{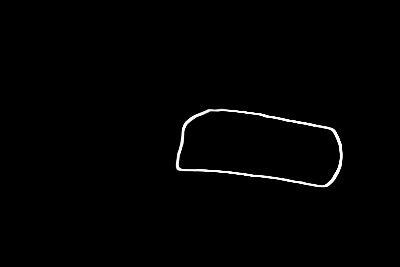}  
\caption{w/ SFE}
\label{fig:wef_ctr}
\end{subfigure}
\begin{subfigure}{.19\linewidth}
\centering
\includegraphics[width=\linewidth]{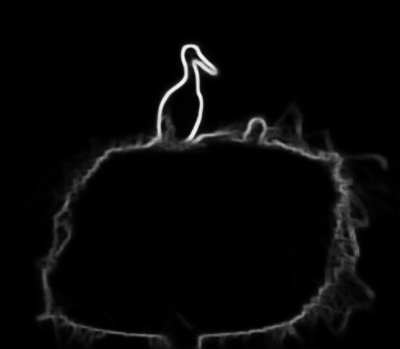}  
\includegraphics[width=\linewidth]{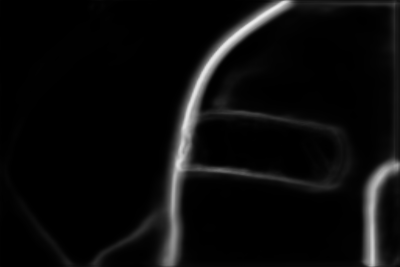}
\caption{w/o SFE}
\label{fig:woef_ctr}
\end{subfigure}
\vspace{-0.03\linewidth}
\caption{Qualitative comparisons for SFE module.  (a) Ground-truth. (b) \& (d) Saliency and contour prediction with SFE module. (c) \& (e) Saliency and contour prediction without SFE module. }
  \label{fig:sfe_module}
\end{figure}

To prove the effectiveness of SFE module, qualitative comparisons are illustrated in Fig.  \ref{fig:sfe_module}. As can be seen, with SFE module, model can effectively suppress execessive wrong predictions. Though SFE is a simple feature extraction module, it improves model predictions greatly.
\vspace{-0.05\linewidth}

More ablation studies please refer to the suplementary material.

\section{Conclusions}
In this paper, we have introduced an efficient and accurate model using a recursive CNN together with a Contour-Saliency Blending (CSB) module. To further improve the model's efficiency, a Stage-wise Feature Extraction (SFE) module is adopted. It is a simple module but can suppress wrong predictions effectively. Furthermore, we divided the training objectives into accuracy and confidence and proposed two loss functions to guide model convergence. The predicted salient objects achieved competitive state-of-the-art results on five benchmark datasets. Detailed ablation studies were conducted, which further proved the model's performance.

{\small
\bibliographystyle{ieee_fullname}
\bibliography{references}
}

\end{document}